\documentclass[accepted]{uai2025} % after acceptance, for a revised version; 
\usepackage[american]{babel}
\usepackage{natbib} % has a nice set of citation styles and commands
\usepackage{mathtools} % amsmath with fixes and additions
\usepackage{booktabs} % commands to create good-looking tables
\usepackage{tikz} % nice language for creating drawings and diagrams

\usepackage{graphicx}

\usepackage{amsmath}
\usepackage{amsthm}
\usepackage{amssymb}

\usepackage[capitalize]{cleveref}
\usepackage{algorithm}
\usepackage{algorithmic}

\usepackage{multirow}
\usepackage{colortbl}

\usepackage{pifont}
\usepackage{xcolor}
\usepackage{xspace}

\newcommand{\R}{\mathbb{R}}
\newcommand{\mymethod}{Contrast-CAT\xspace}
\newcommand{\cmark}{\textcolor{black}{\ding{51}}}%
\newcommand{\xmark}{\textcolor{black}{\ding{55}}}% 55? 53?
\title{Contrast-CAT: Contrasting Activations for Enhanced Interpretability in Transformer-based Text Classifiers}

\author[1]{Sungmin Han}
\author[1]{Jeonghyun Lee}
\author[1]{Sangkyun Lee\thanks{Corresponding author.}}
\affil[1]{%
    School of Cybersecurity, Korea University, Seoul, South Korea
    \texttt{\{sungmin\_15,nomar0107,sangkyun\}@korea.ac.kr}
}

\begin{document}
\maketitle
%\blfootnote{This paper was accepted to the 41st Conference on Uncertainty in Artificial Intelligence (UAI 2025).}
\renewcommand{\thefootnote}{}
\footnotetext{This paper was accepted to the 41st Conference on Uncertainty in Artificial Intelligence (UAI 2025).}
\renewcommand{\thefootnote}{\arabic{footnote}}

\begin{abstract}
Transformers have profoundly influenced AI research, but explaining their decisions remains challenging -- even for relatively simpler tasks such as classification -- which hinders trust and safe deployment in real-world applications. Although activation-based attribution methods effectively explain transformer-based text classification models, our findings reveal that these methods can be undermined by class-irrelevant features within activations, leading to less reliable interpretations. To address this limitation, we propose Contrast-CAT, a novel activation contrast-based attribution method that refines token-level attributions by filtering out class-irrelevant features. By contrasting the activations of an input sequence with reference activations, Contrast-CAT generates clearer and more faithful attribution maps. Experimental results across various datasets and models confirm that Contrast-CAT consistently outperforms state-of-the-art methods. Notably, under the MoRF setting, it achieves average improvements of $\times 1.30$ in AOPC and $\times 2.25$ in LOdds over the most competing methods, demonstrating its effectiveness in enhancing interpretability for transformer-based text classification.
\end{abstract}

\section{Introduction}\label{sec:intro}

\begin{figure}[tb]
\centering
\includegraphics[width=0.435\textwidth]{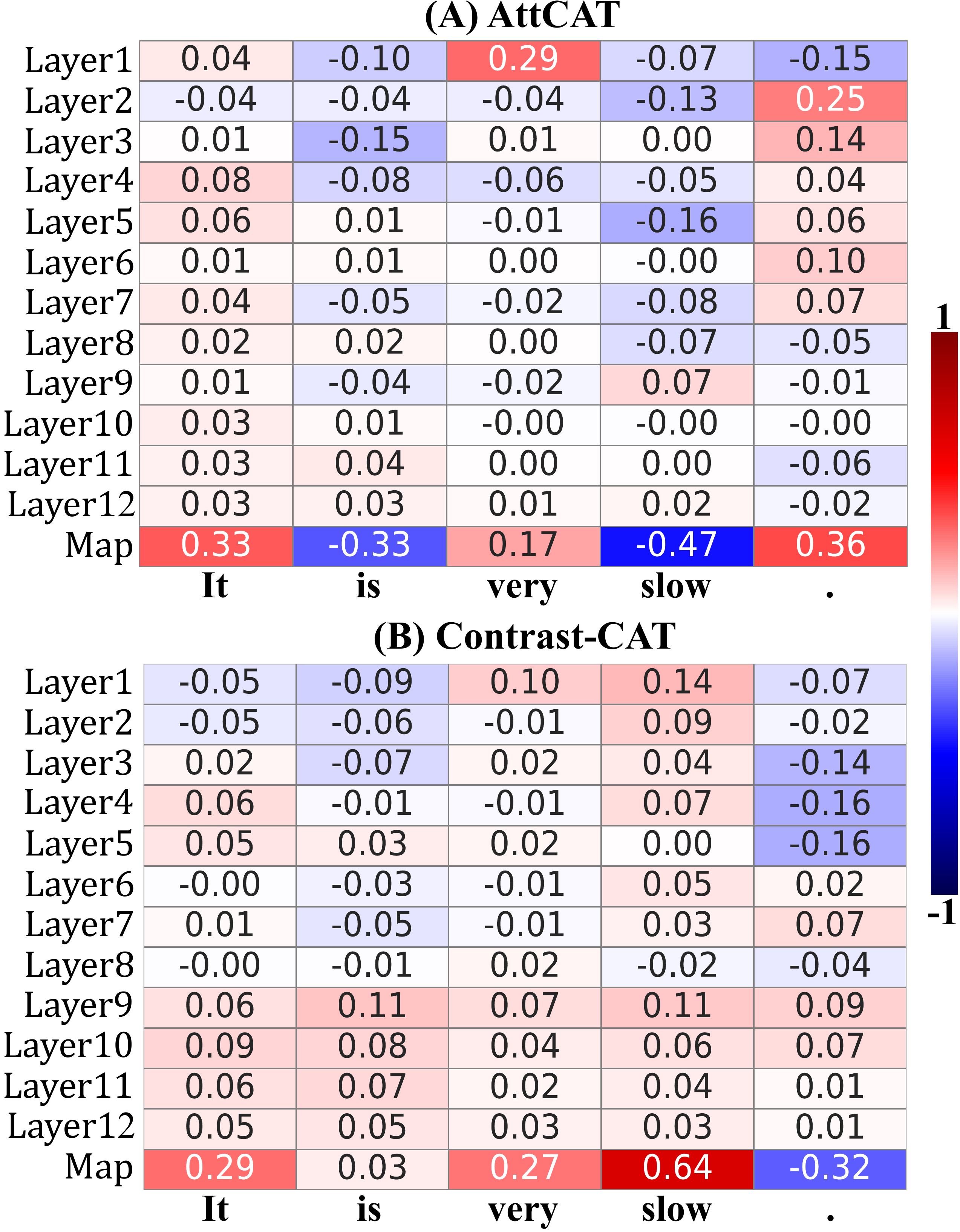}
\caption{Heatmaps displaying attribution values from different encoder layers of the BERT$_{\text{base}}$ model for a negative review prediction. Panel A shows maps generated by AttCAT, which applies gradients directly to activations, while Panel B shows maps from \mymethod, which applies gradients to activation contrast information. Values closer to 1 (red) indicate stronger contributions to the negative prediction.}
\label{fig:motivation}
\end{figure}

Transformers~\citep{transformer} have achieved remarkable success in recent years, transcending both academic and industrial boundaries and becoming increasingly integrated into daily life. However, this widespread integration also heightens the risk of direct exposure to AI errors, underscoring the need to ensure the safety, security, and trustworthiness of AI systems through increased transparency~\citep{biden,NIST_Trustworthy_and_Responsible_AI_100_5,eu_ai_act_2024}. Consequently, developing methods for interpreting the decision-making processes of transformer-based models has become essential.

To address this need, numerous methods have been proposed for interpreting transformer-based models, particularly in text classification tasks where they have shown remarkable performance. These methods generate attribution maps that indicate the relative contributions of input tokens to a model’s decisions. In Section~\ref{sect:related}, we categorize them into attention-based, LRP-based, and activation-based approaches. This work focuses on activation-based attribution, which leverages a model’s activation information to produce attribution maps and has demonstrated state-of-the-art performance in attribution quality.

Activation-based attribution maps are typically derived by extracting activations from one or more layers of a neural network for a given input sequence. Then, the output gradient of the target class, with respect to these activations, is applied to isolate class-relevant features~\citep{GradCAM}. However, we find that this procedure can still be influenced by class-irrelevant signals within the activations, thus limiting its ability to produce accurate, class-specific interpretations.
For example, in \Cref{fig:motivation}, panel (A) illustrates attribution maps generated by AttCAT~\citep{Attcat}, one of the leading activation-based attribution methods, for the movie review `It is very slow.', which is classified as negative. Ideally, the word `slow' should register as highly relevant, with a positive attribution value in relation to the negative sentiment. However, AttCAT fails to detect this importance, whereas our proposed method, \mymethod, correctly assigns the highest attribution to `slow.'

In this paper, we introduce \mymethod, a novel activation-based attribution method for transformer-based text classification. We find that existing methods often incorporate class-irrelevant signals, compromising attribution accuracy. By contrasting target activations with multiple reference activations, \mymethod filters out these irrelevant features and produces high-quality token-level attribution maps. Extensive experiments show that \mymethod consistently outperforms state-of-the-art approaches, achieving average improvements of $\times 1.30$ and $\times 2.25$ in AOPC and LOdds under the MoRF setting, and $\times 1.34$ and $\times 1.03$ under the LeRF setting, compared to the best competitors.

\section{Related Work}\label{sect:related}

We describe attribution methods for interpreting transformer-based text classification models, categorizing them into attention-, LRP-, and activation-based approaches.

\paragraph{Attention-based Attribution}
Attention-based attribution methods rely on attention scores, a key component of transformers.
Under the assumption that input tokens with high attention scores significantly influence model outputs, numerous studies~\citep{attn_xai_2,attn_xai_5,Rollout,globenc,rollout2} have employed attention scores for interpretative purposes of a model.
Specifically, \citet{Rollout} proposed Rollout, which integrates attention scores across multiple layers while accounting for skip connections in transformer architectures to capture information flow.
Additionally, there have been many papers~\citep{attn_grad_xai_2,Gradsam} that introduce the gradient of attention weight for interpretation.
Despite advances in attention-based methods, significant debate remains about whether attention scores truly reflect the relevance of model predictions, as highlighted in~\citep{attention_is_not_explain, attention_not_not_explain}.

\paragraph{LRP-based Attribution}
Layer-wise relevance propagation (LRP)~\citep{LRP_bach} is a technique for backpropagating relevance scores through a neural network, with the scores reflecting our specific interest in the model's prediction.
Building on LRP, several studies have derived explanations for model behavior~\citep{CLRP,PartialLRP,Transatt}.
In~\citep{PartialLRP}, LRP was partially used to determine the most important attention heads within a specific transformer's encoder layer, utilizing relevance scores for the attention weights. 
\citet{Transatt} introduces TransAtt, which propagates relevance scores through all layers of a transformer, combining these scores with gradients of the attention weights and utilizing the Rollout technique for multi-layer integration.
However, LRP-based methods are limited by certain assumptions, known as the LRP rules, designed to uphold the principle of relevance conservation~\citep{lrp_rule}. 

\paragraph{Activation-based Attribution}

In contrast to the methods discussed above, activation-based attribution primarily relies on activation information from each layer of a transformer model. These methods are based on core ideas originally developed for convolutional neural networks (CNNs), which have been shown to be effective for generating high-quality interpretations with simple implementations and broad versatility~\citep{GradCAM,ScoreCAM,activation_noise0,LibraCAM}.
In~\citep{Attcat}, the authors introduced AttCAT as the first adaptation of Grad-CAM~\citep{GradCAM}, one of the most popular activation-based methods for CNNs, to interpret the decisions of transformer-based text classification models. 
AttCAT generates token-level attribution maps by merging activations and their gradients in relation to the model's predictions, following Grad-CAM's essential approach, which uses gradients to reflect class-relevant information.
Similarly, \citet{TIS} introduced TIS adapting Score-CAM~\citep{ScoreCAM}: TIS uses the centroids of activation clusters identified from the activation from all layers to compute relevance scores in a manner akin to Score-CAM.

Although there are attribution methods for transformer-based text classification models that use gradients to extract class-relevant features from activations, no approach has yet focused on filtering out class-irrelevant features through activation contrasting to improve token-level attribution quality.

\section{Preliminary}

\paragraph{Problem Statement}

Consider a pre-trained transformer-based model as a function $f$ processing input tokens $x := \{x_{i}\}_{i=1}^{T}$, where $T$ is the length of the input sequence, and each token is denoted as $x_{i} \in \R^{n}$. 
Our objective is to generate a token-level attribution map $I(x):= \{I(x)_{i}\}_{i=1}^{T}$, where $I(x)_{i}$ represents the relevance score of each input token $x_{i}$ regarding the output $f(x)$.

\paragraph{Transformers}

Let us consider a transformer-based model which is composed of $L$ stacked layers of identical structure. 
We denote that the $\ell$-th layer outputs an activation sequence $A^{\ell}:= \{A^{\ell}_{i}\}_{i=1}^{T}$ that corresponds to input tokens, where $A^{\ell}_{i} \in \R^{n}$.
Each layer computes its output by combining the output from the attention layer with the previous layer's activation, where the attention layer calculates the attention scores:
\begin{equation}\label{eq:transformer.attention}
\begin{aligned}
     \alpha^{\ell,h} := \text{softmax}\left( Q^{\ell,h}(A^{\ell-1}) \cdot K^{\ell,h}(A^{\ell-1})^{T}/\sqrt{d} \right).
\end{aligned}
\end{equation}
Here, $Q^{\ell,h}(\cdot)$, $K^{\ell,h}(\cdot)$, and $V^{\ell,h}(\cdot)$ are the transformations for computing the query, key, and value of the $\ell$-th layer's $h$-th head, respectively, and $d$ is a scaling factor.
$\alpha^{\ell,h} \in \R^{T \times T}$ refers to the attention map of the $h$-th head, which contains attention scores, where $h= 1\dots H$.
We denote by $\tilde A^{\ell,h}$ the output of the $h$-th attention head in the $\ell$-th layer:
\begin{equation*}
\tilde A^{\ell,h} := \alpha^{\ell,h} \cdot V^{\ell,h}(A^{\ell-1}).
\end{equation*}
The outputs from multiple attention heads are concatenated and then combined using a fully connected layer with the skip connection: $\hat A^{\ell} := \text{Concat}(\tilde A^{\ell,1},\tilde A^{\ell,2}, \dots, \tilde A^{\ell,H}) \cdot \tilde W^{\ell} + A^{\ell-1},$
where $\tilde W^{\ell}$ is the weight of the fully connected layer.
Finally, the $\ell$-th layer's output $A^{\ell} \in \R^{T \times n}$ is computed using a feed-forward layer and skip connection:
\begin{equation}\label{eq:transformer.active}
\begin{aligned}
    & A^{\ell} =  \hat A^{\ell} \cdot W^{\ell} + \hat A^{\ell},
\end{aligned}
\end{equation}
where $W^{\ell} \in \R^{n \times n}$ is the weight for the feed-forward layer. 
We have omitted bias parameters and layer normalization in the above expressions for simplicity.

\section{Methodology}\label{sec:mymethod}

\begin{figure*}[tb]
    \centering
    \includegraphics[width=0.89\textwidth]{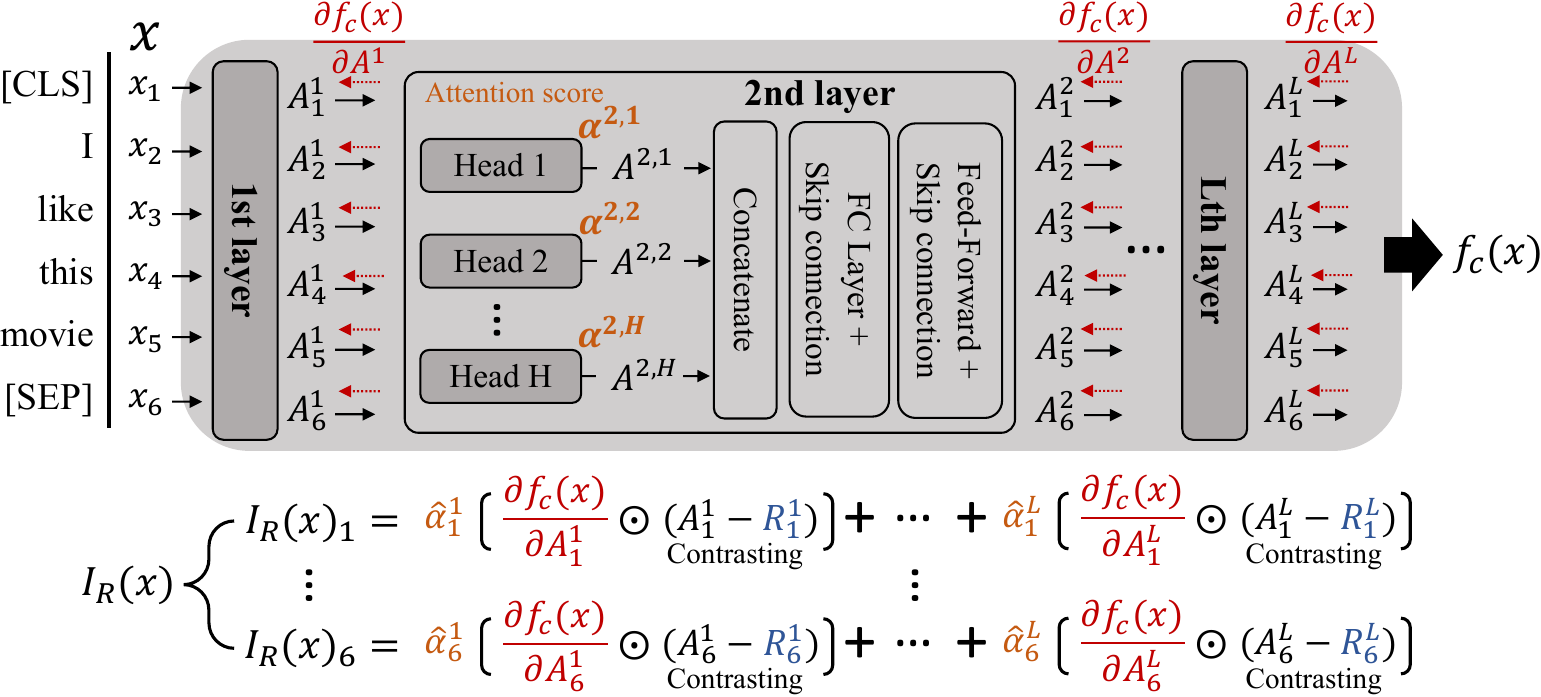}
    \caption{Construction of \mymethod's attribution map. For an input token sequence $x$, \mymethod\ computes an attribution map $I_{R}(x)$ by contrasting the \emph{target activation} $A$ (black) with a \emph{reference activation} $R$ (blue), then weighting by gradients (red) and attention (yellow).}
    \label{fig:overview}
\end{figure*}

We introduce \mymethod, a \emph{token-level}, \emph{activation-based} attribution framework tailored to \emph{transformer} models.

\subsection{Attribution Map}\label{sec:attribution-map}

Let $x := \{x_i\}_{i=1}^T$ be a sequence of $T$ tokens, and let $f_c(x)$ denote the model's score for the target class $c$. For each token $x_i$ ($i=1,\dots,T$), \mymethod\ defines its \emph{attribution} with respect to a \emph{contrastive reference} $R$ as:
\begin{equation}
\label{eq:contrastcat}
I_{R}(x)_{i}
\;:=\;
\sum_{\ell=1}^{L}
\hat{\alpha}^{\ell}_{i}
\sum_{j=1}^{n}
\Bigl(
  \tfrac{\partial f_c(x)}{\partial A^\ell_{i}}
  \;\odot\;
  \bigl(A^\ell_{i} - R^\ell_{i}\bigr)
\Bigr)_{j}.
\end{equation}
Here,
\begin{itemize}
   \item $A^\ell_{i} \in \mathbb{R}^n$ is the activation for token $x_i$ at layer~$\ell$,
   \item $\tfrac{\partial f_c(x)}{\partial A^\ell_{i}} \in \mathbb{R}^n$ is the gradient of $f_c(x)$ w.r.t. $A^\ell_{i}$,
   \item $R^\ell_{i}$ is a \emph{reference activation} for token $i$ chosen from a reference token sequence $r$ such that $f_c(r)<\gamma$,
   \item $\odot$ denotes element-wise multiplication,
   \item $\hat{\alpha}^\ell_i$ is the \emph{averaged attention} of token~$i$ at layer~$\ell$.
\end{itemize}
In essence, $\bigl(A^\ell_i - R^\ell_i\bigr)$ \emph{contrasts} the target activation against one that does not strongly activate class~$c$, thereby removing non-target signals (class-irrelevant features). The factor $\tfrac{\partial f_c(x)}{\partial A^\ell_{i}}$ highlights the parts of the activation that actually affect the model's output, while $\hat{\alpha}^\ell_i$ weights these elements by how much the transformer attends to token~$i$.

Figure~\ref{fig:overview} provides a simplified illustration of the attribution map construction process for \mymethod.

\subsection{Component Details and Motivation}
\label{sec:component_details}

\paragraph{Token-Level Activations $\boldsymbol{A^\ell_i}$.}
Transformers represent each token $x_i$ as a vector in each layer $\ell$. By working at the \emph{token level}, \mymethod\ directly captures the discrete, context-dependent nature of language—differentiating it from CNN-based attribution methods initially designed for spatial feature maps.

\paragraph{Gradients $\boldsymbol{\tfrac{\partial f_c(x)}{\partial A^\ell_i}}$.}
Inspired by gradient-based interpretations, we leverage the partial derivative of $f_c(x)$ w.r.t. $A^\ell_i$. This follows general insights from activation-based methods, (e.g., \citep{GradCAM}), ensuring that only components of $A^\ell_i$ that genuinely influence $f_c(x)$ are emphasized.

\paragraph{Activation Contrasting $\boldsymbol{A^\ell_i - R^\ell_i}$.}
A key novelty of \mymethod\ is its \emph{contrast} operation, which computes the difference between a target activation $A^\ell_i$ and a \emph{low-activation} reference $R^\ell_i$. The reference $R^\ell_i$ is chosen from a sequence $r$ such that $f_c(r)<\gamma$, where $\gamma$ is a pre-defined small positive number ($\gamma >0$).
This choice ensures that the reference activation has a minimal response to the target class $c$ (we set $\gamma=10^{-3}$ in our experiments).
While the use of reference or baseline activations is broadly motivated by prior works (e.g., \citep{LibraCAM}), \mymethod\ is the first to extend this idea to transformer-based text classification networks, applying it \emph{across multiple transformer layers}, at the \emph{token level}, explicitly targeting textual data. This operation highlights class-specific features that distinguish $x$ from a weakly activating example.

\paragraph{Attention Weights $\boldsymbol{\hat{\alpha}^\ell_i}$.}
Transformers distribute relevance across tokens via multi-head attention. We aggregate these attention scores into $\hat{\alpha}^\ell_i$, giving higher importance to tokens that the model itself regards as salient. Unlike purely attention-based methods (e.g., \citep{Rollout}), \mymethod\ integrates attention and gradient-based cues, offering a more robust attribution signal.

\paragraph{Multi-Layer Attribution}
Building on prior findings that transformers encode varying levels of semantic information across their layers---ranging from phrase-level details to deeper semantic features~\citep{level_of_semantics3,level_of_semantics1,emnlp_review_telling_bert}---we diverge from traditional activation-based attribution methods which typically rely on a single layer (e.g., \citep{Gradsam}). Instead, we incorporate \emph{multi-layer} activations $A^\ell$ from all layers $\ell = 1, \dots, L$ in \cref{eq:transformer.active}, together with their layer-wise attention scores $\alpha^{\ell,h}$ in \cref{eq:transformer.attention}. This design captures \emph{layer-specific} token semantics, and by weighting them with $\hat{\alpha}^\ell_i$, it effectively highlights the tokens most influential to the model's output across all layers.

\subsection{Attribution with Multiple Contrast}
\label{sec:multi-references}

Relying on a \emph{single} reference from one class can be insufficient if the target activations $A^\ell := \{A^{\ell}_{i}\}_{i=1}^{T}$ encode features shared across \emph{multiple} non-target classes. Moreover, any features that consistently remain after contrasting $A^\ell$ with several reference activations are more likely to represent class-specific properties. To address this, we generate a collection of attribution maps 
\begin{align*}
    D \;:=\;
  \bigl\{
    I_{R(r)}(x)
    \;\bigm|\;
    r \in \text{training set},\;
    f_{c}(r) < \gamma
  \bigr\},
\end{align*}
by repeating the procedure in Section~\ref{sec:attribution-map} with \emph{multiple} reference sequences. We cache these reference activations---one might call it a \emph{reference library}---for use during inference. In practice, we employ $30$ pre-computed references per class.

\paragraph{Refinement via Deletion Test}
\label{method.selectively_fitering_multicont}
Although this multi-reference approach reduces the risk of overlooking crucial class-relevant features, not all resulting maps $I_{R(r)}(x)$ are guaranteed to be reliable. We therefore \emph{refine} \mymethod by examining each map’s \emph{attribution quality} using a token-wise deletion test~(e.g., \citep{deletion,ScoreCAM}). Specifically, we remove the top-attributed tokens one by one and record how much the model’s predictive probability for class~$c$ decreases. The \emph{average probability drop score} captures, on a token-by-token basis, how effectively a map localizes truly important tokens.  

Any map with a drop score below a specified threshold $\rho$ (set in our experiments to the mean plus one standard deviation of all drop scores) is discarded. Finally, we generate the \mymethod\ attribution by averaging all remaining high-quality maps:
\begin{align*}
  &I(x) 
  \;:=\; 
  \frac{1}{|M|}
  \sum_{I_{R}(x)\,\in\,M}
  I_{R}(x),
  \\
  &\text{where}
  \quad
  M := \bigl\{\, I_{R}(x) \in D : S\bigl(I_{R}(x)\bigr) \ge \rho \bigr\}.
\end{align*}
This final aggregation fuses the most credible contrastive perspectives into a single, robust token-level attribution.  

\section{Experiments}

\paragraph{Experiment Settings}
We implemented our method, Contrast-CAT, using PyTorch (the code is available at \url{https://github.com/ku-air/Contrast-CAT}).
We used the BERT$_{\text{base}}$ model~\citep{bert}, consisting of $12$ encoder layers with $12$ attention heads, as the transformer-based model for our experiments (see the supplementary material for results using other transformer-based models). We evaluated our method on four popular datasets for text classification tasks: Amazon Polarity~\citep{amazon_yelp}, Yelp Polarity~\citep{amazon_yelp}, SST2~\citep{sst}, and IMDB~\citep{imdb}.
We reported our results using $2000$ random samples from the test sets of each dataset, except for SST2, for which the entire set was used since the entire dataset had fewer than $2000$ samples.

We compared our method to various attribution methods, categorized by attention-based: RawAtt, Rollout~\citep{Rollout}, Att-grads, Att$\times$Att-grads, and Grad-SAM~\citep{Gradsam}; LRP-based: Full LRP~\citep{FLRP}, Partial LRP~\citep{PartialLRP}, and TransAtt~\citep{Transatt}; and activation-based methods: CAT, AttCAT~\citep{Attcat}, and TIS~\citep{TIS}.

\paragraph{Evaluation Metrics}

We used the area over the perturbation curve (denoted by AOPC)~\citep{aopc1,aopc2_lodss2} and the log-odds (LOdds)~\citep{lodss,aopc2_lodss2} metrics for assessing the faithfulness of attribution following the previous research~\citep{Attcat}. Faithfulness refers to the accuracy with which an attribution map's scores reflect the actual influence of each input token on the model's prediction. The AOPC and LOdds metrics are defined as follows: (1) AOPC($k$) := $\frac{1}{N}\sum_{i=1}^{N} (y_i^c - \tilde y_i^c)$, and (2) LOdds($k$) := $\frac{1}{N}\sum_{i=1}^{N} \log\left(\frac{\tilde y_i^c}{y_i^c}\right)$.
Here, $N$ is the total number of data points used for evaluation, and $y_i^c$ denotes the model's prediction probability for the class $c$ of a given input token sequence $x$, while $\tilde y_i^c$ indicates the probability after removing the top-$k\%$ of input tokens based on relevance scores from an attribution map.

To evaluate attribution quality more precisely using the AOPC and LOdds metrics while addressing inconsistencies from token removal order (i.e., removing the most relevant tokens first versus the least relevant tokens first)~\citep{morf_lerf_consistency}, we conducted experiments under two settings: one where tokens were removed in descending order of relevance scores (MoRF: Most Relevant First), and another in ascending order (LeRF: Least Relevant First).
Consistently achieving high-quality attribution under both conditions indicates superior attribution quality. 
Specifically, under the MoRF setting, higher AOPC and lower LOdds indicate better attribution, while under the LeRF setting, lower AOPC and higher LOdds suggest better performance.

\subsection{Faithfulness of Attribution}\label{exp.faithful}

\begin{figure*}[tb]
\centering
\includegraphics[width=0.93\textwidth]{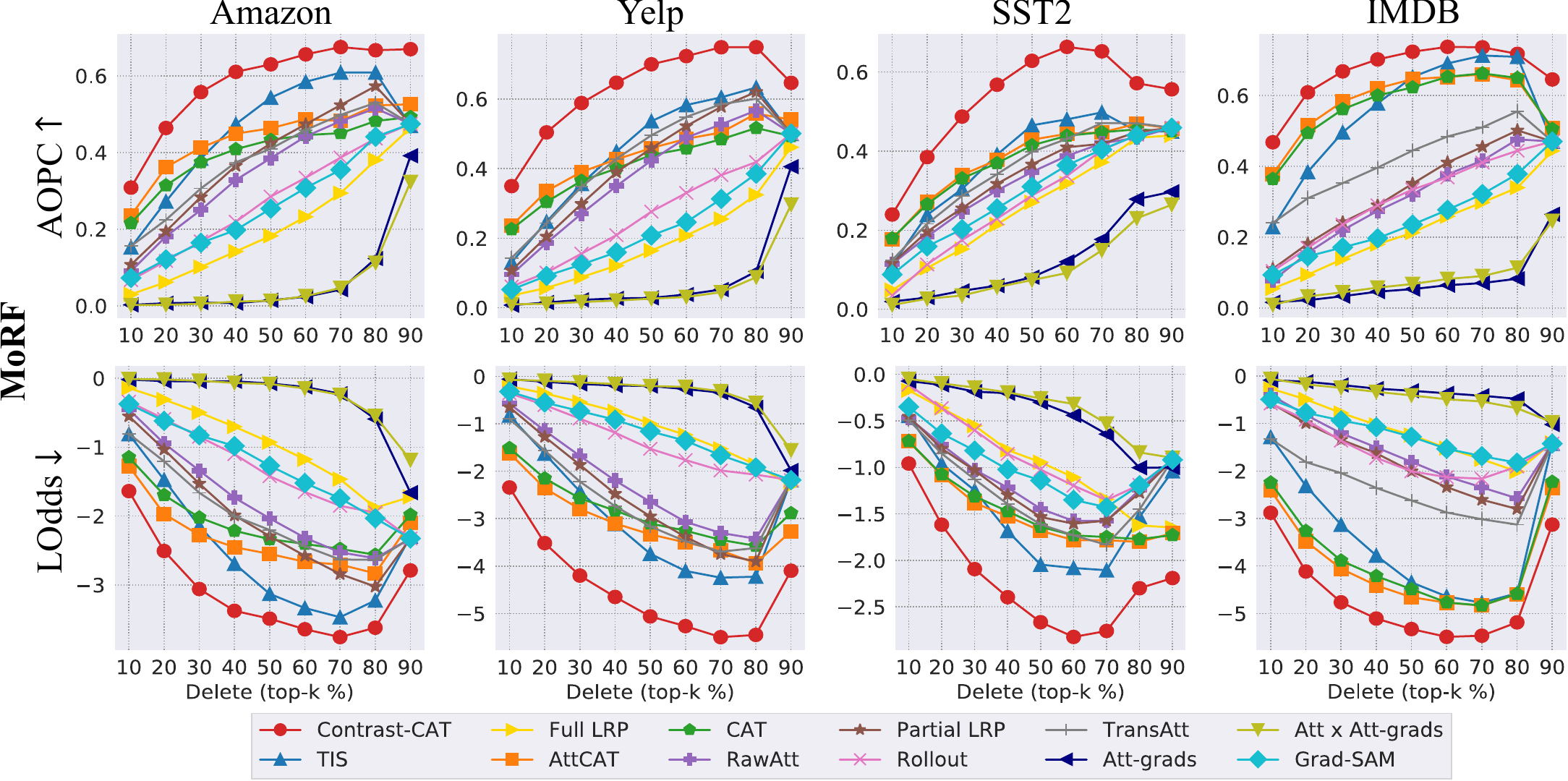}
\caption{Quantitative comparison of the faithfulness evaluation of \mymethod and other attribution methods, measured under the MoRF (Most Relevant First) setting. The arrows mean that $\uparrow$: higher is better, and $\downarrow$: lower is better.}
\label{fig:plot.quantitative}
\end{figure*}

\begin{table*}[tb]
\centering
\caption{AUC values from the faithfulness evaluation, with (A) showing results under the MoRF (Most Relevant First) setting and (B) showing results under the LeRF (Least Relevant First) setting. The best and second-best results are highlighted in bold and underlined, respectively. The arrows mean that $\uparrow$: higher is better, and $\downarrow$: lower is better.}\label{tbl:all.measures}

\newcolumntype{C}[1]{>{\centering\arraybackslash}m{#1}}
\begin{tabular}{c|c|c|c|c|c|c|c|c}\toprule
\multicolumn{9}{c}{(A) MoRF (Most Relevant First)} \\\midrule
 Dataset& \multicolumn{2}{c|}{Amazon} & \multicolumn{2}{c|}{Yelp} & \multicolumn{2}{c|}{SST2}& \multicolumn{2}{c}{IMDB} \\\midrule%\cline{2-7}
 Method & AOPC$\uparrow$ & LOdds$\downarrow$ & AOPC$\uparrow$ & LOdds$\downarrow$ & AOPC$\uparrow$ & LOdds$\downarrow$ & AOPC$\uparrow$ & LOdds$\downarrow$ \\
\midrule

RawAtt & 0.424 & 0.405 & 0.412 & 0.462 & 0.386 & 0.471 & 0.335 & 0.564 \\ 
Rollout & 0.327 & 0.516 & 0.282 & 0.601 & 0.329 & 0.558 & 0.339 & 0.566 \\
Att-grads & 0.061 & 0.749 & 0.059 & 0.754 & 0.132 & 0.691 & 0.061 & 0.759 \\
Att$\times$Att-grads & 0.054 & 0.756 & 0.045 & 0.763 & 0.109 & 0.711 & 0.075 & 0.746 \\
Grad-SAM & 0.312 & 0.526 & 0.235 & 0.633 & 0.356 & 0.518 & 0.266 & 0.623 \\
Full LRP & 0.242 & 0.592 & 0.190 & 0.652 & 0.310 & 0.538 & 0.233 & 0.631 \\
Partial LRP & 0.463 & 0.356 & 0.447 & 0.422 & 0.400 & 0.461 & 0.364 & 0.538 \\
TransAtt & 0.461 & 0.366 & 0.473 & 0.404 & 0.432 & 0.428 & 0.458 & 0.455 \\
CAT & 0.482 & 0.341 & 0.440 & 0.383 & 0.452 & 0.382 & 0.632 & 0.215 \\
AttCAT & 0.527 & 0.292 & 0.470 & \underline{0.346} & 0.461 & 0.372 & \underline{0.644} & \underline{0.198} \\
TIS & \underline{0.560} & \underline{0.241} & \underline{0.494} & 0.349 & \underline{0.463} & \underline{0.367} & 0.618 & 0.277 \\
\mymethod & \textbf{0.703} & \textbf{0.117} & \textbf{0.687} & \textbf{0.131} & \textbf{0.654} & \textbf{0.157} & \textbf{0.738} & \textbf{0.101}
\\\toprule
\multicolumn{9}{c}{(B) LeRF (Least Relevant First)} \\\midrule
Dataset& \multicolumn{2}{c|}{Amazon} & \multicolumn{2}{c|}{Yelp} & \multicolumn{2}{c|}{SST2}& \multicolumn{2}{c}{IMDB} \\\midrule%\cline{2-7}
 Method & AOPC$\downarrow$ & LOdds$\uparrow$ & AOPC$\downarrow$ & LOdds$\uparrow$ & AOPC$\downarrow$ & LOdds$\uparrow$ & AOPC$\downarrow$ & LOdds$\uparrow$ \\
\midrule

RawAtt & 0.133 & 0.694 & 0.093 & 0.723 & 0.249 & 0.577 & 0.158 & 0.688 \\ 
Rollout & 0.166 & 0.670 & 0.130 & 0.687 & 0.373 & 0.448 & 0.126 & 0.711 \\
Att-grads & 0.636 & 0.186 & 0.560 & 0.252 & 0.601 & 0.223 & 0.588 & 0.271 \\
Att$\times$Att-grads & 0.707 & 0.111 & 0.660 & 0.145 & 0.681 & 0.126 & 0.709 & 0.127 \\
Grad-SAM & 0.139 & 0.677 & 0.107 & 0.713 & 0.285 & 0.547 & 0.118 & 0.715 \\
Full LRP & 0.254 & 0.588 & 0.187 & 0.649 & 0.377 & 0.454 & 0.199 & 0.656 \\
Partial LRP & 0.122 & 0.700 & 0.088 & 0.725 & 0.237 & 0.585 & 0.134 & 0.701 \\
TransAtt & 0.089 & 0.731 & \underline{0.063} & \underline{0.751} & 0.215 & 0.605 & \underline{0.061} & \underline{0.761} \\
CAT & 0.108 & 0.712 & 0.087 & 0.727 & 0.213 & 0.611 & 0.128 & 0.697 \\
AttCAT & \underline{0.078} & \underline{0.740} & \underline{0.063} & 0.747 & \underline{0.205} & \underline{0.623} & 0.119 & 0.703 \\
TIS & 0.104 & 0.719 & 0.082 & 0.737 & 0.252 & 0.562 & 0.135 & 0.691 \\
\mymethod & \textbf{0.058} & \textbf{0.757} & \textbf{0.048} & \textbf{0.759} & \textbf{0.147} & \textbf{0.669} & \textbf{0.047} & \textbf{0.775}
\\\bottomrule
\end{tabular}
\end{table*}

Figure~\ref{fig:plot.quantitative} illustrates the AOPC and LOdds values for attribution maps generated by each competing method, evaluated at various top-$k\%$ thresholds where $k$ is increased by $10$ within the range of $[10,90]$. Table~\ref{tbl:all.measures} provides the corresponding AUC values. 
Note that Figure~\ref{fig:plot.quantitative} presents results for the MoRF setting only, while Table~\ref{tbl:all.measures} includes results for both MoRF and LeRF settings.
Through this evaluation, we can analyze the overall characteristics of an attribution map in terms of relevance scores of different threshold levels.

The trends in Figure~\ref{fig:plot.quantitative} reveal that our method, \mymethod, consistently maintains faithful attribution quality across all threshold levels and datasets compared to other methods. Table~\ref{tbl:all.measures} further supports this, showing that \mymethod achieves top-$1$ attribution quality under both MoRF and LeRF settings. Specifically, compared to the second-best cases, \mymethod shows average improvements in AUC values of AOPC and LOdds under the MoRF setting by $\times 1.30$ and $\times 2.25$, respectively. For the LeRF setting, \mymethod shows average improvements in AUC values of AOPC and LOdds by $\times 1.34$ and $\times 1.03$, respectively.

\subsection{Qualitative Evaluation}
\begin{figure*}[tb]
\centering
\includegraphics[width=0.93\textwidth]{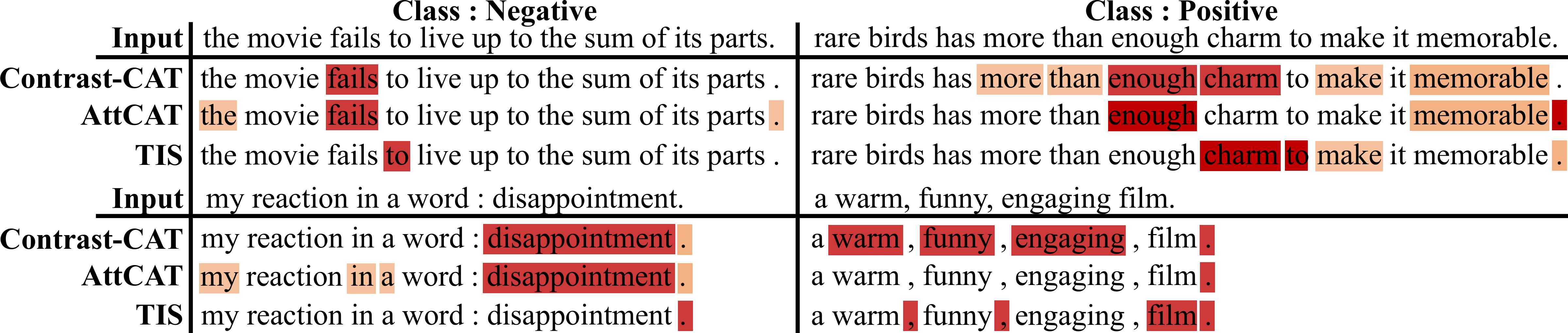}
\caption{Qualitative comparison of attribution quality. Relevance scores are shown with color shades: red for the highest importance, followed by orange.}
\label{fig:quality}
\end{figure*}

Figure~\ref{fig:quality} illustrates the attribution maps generated by \mymethod, TIS, and AttCAT, the top-$3$ ranked methods in our faithfulness evaluation, conducted under the MoRF setting (Table~\ref{tbl:all.measures}, (A) MoRF).
The examples provided are from the SST2 dataset. For ease of interpretation, only tokens with relevance scores exceeding $0.5$ are highlighted. 
As shown in the left side of Figure~\ref{fig:quality}, \mymethod identifies relevant tokens related to the predicted class, such as `fails' or `disappointment' for the negative prediction cases.
For a positive prediction, in the input phrase `rare birds have more than enough charm to make it memorable.', \mymethod highlights `enough' and `charm' as the most relevant tokens, with `than', `make', `more', and `memorable' following in relevance. In contrast, AttCAT focuses only on `enough' and `memorable', missing `charm' and `more', while TIS identifies `to' as the most relevant token.

\subsection{The Effect of Activation Contrast}

\begin{table*}[tb]
\centering
\caption{The effect of our activation contrasting approach, measured under the MoRF (Most Relevant First) setting. `Random' uses randomly selected references (the mean values over $30$ repetitions are reported), `Same' uses references from the same class as the target, and `Contrasting' refers to the suggested \mymethod. The best results are in boldface.}\label{tbl:ablation.denoise_effect}
\newcolumntype{C}[1]{>{\centering\arraybackslash}m{#1}}
\begin{tabular}{c|c|c|c|c|c|c|c|c}\toprule
 Dataset& \multicolumn{2}{c|}{Amazon} & \multicolumn{2}{c|}{Yelp} & \multicolumn{2}{c|}{SST2}& \multicolumn{2}{c}{IMDB} \\\midrule%\cline{2-7}
 Reference & AOPC$\uparrow$ & LOdds$\downarrow$ & AOPC$\uparrow$ & LOdds$\downarrow$ & AOPC$\uparrow$ & LOdds$\downarrow$ & AOPC$\uparrow$ & LOdds$\downarrow$ \\
\midrule

Random & 0.513 & 0.306 & 0.496 & 0.323 & 0.433 & 0.398 & 0.634 & 0.213 \\
Same & 0.144 & 0.667 & 0.159 & 0.650 & 0.089 & 0.728 & 0.124 & 0.614 \\
Contrasting & \textbf{0.703} & \textbf{0.117} & \textbf{0.687} & \textbf{0.131} & \textbf{0.654} & \textbf{0.157} & \textbf{0.738} & \textbf{0.101} \\\bottomrule
\end{tabular}
\end{table*}

To evaluate the effect of our \mymethod's activation contrasting, we compared the attribution quality of different versions of \mymethod: the `Random' version uses randomly selected references from individual training datasets instead of what had been outlined in Section~\ref{sec:multi-references}, and the `Same' version uses references of the same class as the target instead of different classes.
The `Same' version contrasts with our method, which leverages activations from different classes as contrastive references.

Table~\ref{tbl:ablation.denoise_effect} presents AUC values of each version of \mymethod, where the suggested \mymethod is denoted by `Contrasting'.
The attribution quality is the worst with `Same' and the best with `Contrasting', which indicates that the proposed activation contrasting effectively reduces non-target signals in the activations, thereby helping to generate high-quality attribution maps.

\subsection{Confidence of Attribution}

\begin{table}[tb]
\centering
\caption{The results of the confidence evaluation, showing averaged rank correlation values. The values below $0.05$ (marked in gray) indicate that attributions tend to be class-distinct, as desired.}\label{tbl:all.k_tau}
\setlength{\tabcolsep}{2pt}
\begin{tabular}{c|c|c|c|c}\toprule
 \multirow{2}{*}{Method} & \multicolumn{4}{c}{Dataset} \\\cline{2-5}
  & Amazon & Yelp & SST2 & IMDB \\
\midrule

RawAtt & 1.00 & 1.00 & 1.00 & 1.00 \\ 
Rollout & 1.00 & 1.00 & 1.00 & 1.00 \\ 
Att-grads & \cellcolor[gray]{0.8}$<$ 0.05 & \cellcolor[gray]{0.8}$<$ 0.05 & \cellcolor[gray]{0.8}$<$ 0.05 & \cellcolor[gray]{0.8}$<$ 0.05 \\ 
Att$\times$Att-grads & \cellcolor[gray]{0.8}$<$ 0.05 & \cellcolor[gray]{0.8}$<$ 0.05 & \cellcolor[gray]{0.8}$<$ 0.05 & \cellcolor[gray]{0.8}$<$ 0.05 \\ 
Grad-SAM & 0.158 & 0.138 & 0.282 & 0.084 \\ 
Full LRP & 0.732 & 0.629 & 0.712 & 0.533 \\ 
Partial LRP & 0.952 & 0.924 & 0.957 & 0.859 \\ 
TransAtt & 0.153 & 0.135 & 0.342 & 0.061 \\ 
CAT & \cellcolor[gray]{0.8}$<$ 0.05 & \cellcolor[gray]{0.8}$<$ 0.05 & \cellcolor[gray]{0.8}$<$ 0.05 & \cellcolor[gray]{0.8}$<$ 0.05 \\ 
AttCAT & \cellcolor[gray]{0.8}$<$ 0.05 & \cellcolor[gray]{0.8}$<$ 0.05 & \cellcolor[gray]{0.8}$<$ 0.05 & \cellcolor[gray]{0.8}$<$ 0.05 \\ 
TIS & \cellcolor[gray]{0.8}$<$ 0.05 & \cellcolor[gray]{0.8}$<$ 0.05 & \cellcolor[gray]{0.8}$<$ 0.05 & \cellcolor[gray]{0.8}$<$ 0.05 \\ 
\mymethod & \cellcolor[gray]{0.8}$<$ 0.05 & \cellcolor[gray]{0.8}$<$ 0.05 & \cellcolor[gray]{0.8}$<$ 0.05 & \cellcolor[gray]{0.8}$<$ 0.05 \\ 

\bottomrule
\end{tabular}
\end{table}

If an attribution method consistently generates similar attribution maps regardless of the model's prediction, its confidence is questionable. Therefore, we conducted the confidence evaluation of the attribution methods employing the Kendall-$\tau$ rank correlation~\citep{kendall1948rank}, which is a statistical measure used to assess the similarity between two data by comparing the ranking order of their respective values. We compute an averaged rank correlation:
\begin{equation*}
\frac{1}{N}\sum_{i=1}^{N} \text{Kendall-}\tau(P^{c}_{i}, P^{\hat c}_{i}),
\end{equation*}
where $P^{c}_{i}$ is an array of token indices in descending order of relevance scores for class $c$ in an attribution map, $P^{\hat c}_{i}$ is a similar array but for the class $\hat c \neq c$, and $N$ is the total number of data points used for testing.
For the choice of $\hat c$, we followed the settings of AttCAT as detailed in their open-source implementation, where the class immediately following the class $c$ was chosen.

If an attribution method assigns relevance scores to tokens in distinct orders for different class predictions of the inspected model, the rank correlation is expected to be low.
Table~\ref{tbl:all.k_tau} presents the average rank correlation for various attribution methods tested across datasets. Cases with average rank correlation values under $0.05$ are marked as `$<0.05$' and highlighted: these are the cases where the attribution methods seem to work soundly -- our \mymethod seems to pass the test, along with Att-grads, Att$\times$Att-grads, CAT, AttCAT and TIS.
In contrast, methods such as RawAtt, Rollout, and Partial LRP showed values near $1.0$ consistently over the datasets, suggesting that these methods have issues generating distinct attribution over different class outcomes.

\subsection{The Effect of Using Multiple Layers}

\begin{figure*}[tb]
\centering
\includegraphics[width=0.93\textwidth]{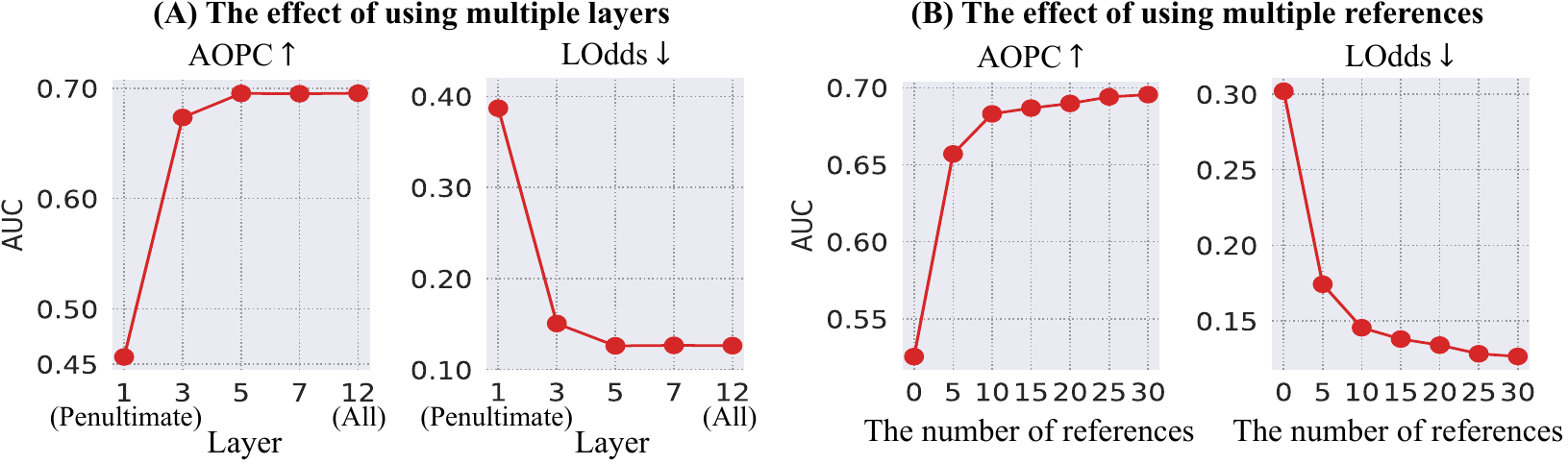}
\caption{Comparison of \mymethod's attribution quality measured under the MoRF (Most Relevant First) setting: (A) varying the number of layers from penultimate to all, and (B) varying the number of reference samples from $0$ to $30$.}
\label{fig:plot.l_wise}
\end{figure*}

Panel (A) of Figure~\ref{fig:plot.l_wise} demonstrates the effect of using multiple layers to improve the attribution quality of \mymethod. The figure shows the average AUC values of AOPC and LOdds across datasets, measured under the MoRF setting.

The results in panel (A) of Figure~\ref{fig:plot.l_wise} indicate that the attribution quality improves as the number of layers increases, with the best performance achieved when all layers are used. 
Specifically, there is a $\times 1.52$ improvement in AOPC and $\times 3.05$ improvement in LOdds when using all layers compared to using only the penultimate layer. The AOPC and LOdds values tend to saturate when we use three or more layers but continue to increase as the number increases. 

\subsection{The Effect of Multiple Contrasts}

Panel (B) of Figure~\ref{fig:plot.l_wise} illustrates the impact of increasing the number of references for multiple contrasts in \mymethod on attribution quality, measured by average AUC for AOPC and LOdds across datasets under the MoRF setting.

The AOPC metric shows a sharp improvement as the number of references increases from $0$ to $5$. After $5$ references, the AUC continues to increase, stabilizing between $25$ and $30$ references. In contrast, the LOdds metric exhibits a sharp decline as the number of references increases, starting at approximately $0.30$ and dropping steadily, stabilizing around $0.10$ after $10$ references and reaching its minimum at $30$ references.
These results indicate that more references improve attribution quality, with the best performance at $30$, which we use in our experiments.

\subsection{The Effect of Contrasting References}

\begin{table}[tb]
\centering
\caption{Impact of the parameter $\gamma$ in the condition \(f_c(r)<\gamma\) on the attribution quality of \mymethod.}\label{tbl:gamma}
\begin{tabular}{c|c|c|c}\toprule
  $\gamma$ & 0.1 & 0.01 & 0.001 \\
\midrule

AOPC$\uparrow$ & 0.627 & 0.651 & 0.696  \\ 
LOdds$\downarrow$ & 0.450 & 0.448 & 0.127 \\ 

\bottomrule
\end{tabular}
\end{table} 

Table~\ref{tbl:gamma} presents the impact of the parameter $\gamma$ in the condition for selecting contrastive references, \(f_c(r)<\gamma\), on \mymethod's attribution quality.
This condition ensures that selected reference activations do not strongly respond to the target class $c$, thereby helping to reduce non-target signals within the target activation by contrasting it with the selected reference activations.

We evaluated \mymethod's faithfulness by varying $\gamma$ from $0.1$ to $0.001$, and reported average AUC values for AOPC and LOdds across datasets under the MoRF setting.
The results in Table~\ref{tbl:gamma} indicate that a smaller $\gamma$ improves \mymethod's attribution quality, highlighting the benefits of low-activation references for activation contrasting, as described in Section~\ref{sec:component_details}.

\section{Conclusion}

In this work, we introduced \mymethod, a novel activation-based attribution method that leverages activation contrasting to generate high-quality token-level attribution map.
Our extensive experiments demonstrated that \mymethod significantly outperforms state-of-the-art methods across various datasets and models.

Despite its effectiveness, \mymethod requires reference points whose activations will be available during the creation of attribution maps.
While we minimized overhead with a pre-built reference library, its storage requirements grow with the number of classes and activation size. Future work will explore lower-cost alternative tensors.

As the demand for interpretable AI grows to support safety, security, and trustworthiness, we believe \mymethod represents a meaningful step toward improving the transparency of transformer-based models.

\begin{acknowledgements}
This work was supported by the Institute of Information \& Communications Technology Planning \& Evaluation(IITP) grant funded by the Korea government(MSIT) (RS-2024-00439819, AI-Based Automated Vulnerability Detection and Safe Code Generation) and by the IITP-ITRC(Information Technology Research Center) grant funded by the Korea government(MSIT)(IITP-2025-RS-2020-II201749).
\end{acknowledgements}

% References
\bibliography{uai2025_arXiv}

\begin{thebibliography}{46}
\providecommand{\natexlab}[1]{#1}
\providecommand{\url}[1]{\texttt{#1}}
\expandafter\ifx\csname urlstyle\endcsname\relax
  \providecommand{\doi}[1]{doi: #1}\else
  \providecommand{\doi}{doi: \begingroup \urlstyle{rm}\Url}\fi

\bibitem[Abnar and Zuidema(2020)]{Rollout}
Samira Abnar and Willem Zuidema.
\newblock Quantifying attention flow in transformers.
\newblock In \emph{ACL}, 2020.

\bibitem[Bach et~al.(2015)Bach, Binder, Montavon, Klauschen, M{\"u}ller, and Samek]{LRP_bach}
Sebastian Bach, Alexander Binder, Gr{\'e}goire Montavon, Frederick Klauschen, Klaus-Robert M{\"u}ller, and Wojciech Samek.
\newblock On pixel-wise explanations for non-linear classifier decisions by layer-wise relevance propagation.
\newblock \emph{PLOS ONE}, 2015.

\bibitem[Barkan et~al.(2021)Barkan, Hauon, Caciularu, Katz, Malkiel, Armstrong, and Koenigstein]{Gradsam}
Oren Barkan, Edan Hauon, Avi Caciularu, Ori Katz, Itzik Malkiel, Omri Armstrong, and Noam Koenigstein.
\newblock Grad-sam: Explaining transformers via gradient self-attention maps.
\newblock In \emph{CIKM}, 2021.

\bibitem[Chefer et~al.(2021)Chefer, Gur, and Wolf]{Transatt}
Hila Chefer, Shir Gur, and Lior Wolf.
\newblock Transformer interpretability beyond attention visualization.
\newblock In \emph{CVPR}, 2021.

\bibitem[Chen et~al.(2020)Chen, Zheng, and Ji]{aopc2_lodss2}
Hanjie Chen, Guangtao Zheng, and Yangfeng Ji.
\newblock Generating hierarchical explanations on text classification via feature interaction detection.
\newblock In \emph{ACL}, 2020.

\bibitem[Chrysostomou and Aletras(2021)]{attn_grad_xai_2}
George Chrysostomou and Nikolaos Aletras.
\newblock Enjoy the salience: Towards better transformer-based faithful explanations with word salience.
\newblock In \emph{EMNLP}, 2021.

\bibitem[Clark et~al.(2019)Clark, Khandelwal, Levy, and Manning]{attn_xai_5}
Kevin Clark, Urvashi Khandelwal, Omer Levy, and Christopher~D. Manning.
\newblock What does {BERT} look at? {An} analysis of {BERT}{'}s attention.
\newblock In \emph{ACL (workshop)}, 2019.

\bibitem[Del~Corso et~al.(2005)Del~Corso, Gull\'{\i}, and Romani]{agnews}
Gianna~M. Del~Corso, Antonio Gull\'{\i}, and Francesco Romani.
\newblock Ranking a stream of news.
\newblock In \emph{WWW}, 2005.

\bibitem[Devlin et~al.(2019)Devlin, Chang, Lee, and Toutanova]{bert}
Jacob Devlin, Ming-Wei Chang, Kenton Lee, and Kristina Toutanova.
\newblock {BERT}: Pre-training of deep bidirectional transformers for language understanding.
\newblock In \emph{NAACL-HLT}, 2019.

\bibitem[Ding et~al.(2017)Ding, Liu, Luan, and Sun]{FLRP}
Yanzhuo Ding, Yang Liu, Huanbo Luan, and Maosong Sun.
\newblock Visualizing and understanding neural machine translation.
\newblock In \emph{ACL}, 2017.

\bibitem[Dunietz et~al.(2024)Dunietz, Tabassi, Latonero, and Roberts]{NIST_Trustworthy_and_Responsible_AI_100_5}
Jesse Dunietz, Elham Tabassi, Mark Latonero, and Kamie Roberts.
\newblock A plan for global engagement on ai standards.
\newblock NIST Trustworthy and Responsible AI, National Institute of Standards and Technology, Gaithersburg, MD, 2024.
\newblock URL \url{https://tsapps.nist.gov/publication/get_pdf.cfm?pub_id=958389}.

\bibitem[Englebert et~al.(2023)Englebert, Stassin, Nanfack, Mahmoudi, Siebert, Cornu, and De~Vleeschouwer]{TIS}
Alexandre Englebert, S\'edrick Stassin, G\'eraldin Nanfack, Sidi~Ahmed Mahmoudi, Xavier Siebert, Olivier Cornu, and Christophe De~Vleeschouwer.
\newblock Explaining through transformer input sampling.
\newblock In \emph{ICCV (workshop)}, 2023.

\bibitem[{European Commission}(2024)]{eu_ai_act_2024}
{European Commission}.
\newblock Artificial intelligence act (regulation (eu) 2024/1689).
\newblock Official Journal of the European Union, 2024.
\newblock URL \url{https://artificialintelligenceact.eu/the-act/}.

\bibitem[F.R.S.(1901)]{pca}
Karl~Pearson F.R.S.
\newblock Liii. on lines and planes of closest fit to systems of points in space.
\newblock \emph{The London, Edinburgh, and Dublin Philosophical Magazine and Journal of Science}, 1901.

\bibitem[Gu et~al.(2018)Gu, Yang, and Tresp]{CLRP}
Jindong Gu, Yinchong Yang, and Volker Tresp.
\newblock Understanding individual decisions of cnns via contrastive backpropagation.
\newblock In \emph{ACCV}, 2018.

\bibitem[Han et~al.(2022)Han, Lee, and Lee]{activation_noise0}
Sungmin Han, Jeonghyun Lee, and Sangkyun Lee.
\newblock Activation fine-tuning of convolutional neural networks for improved input attribution based on class activation maps.
\newblock \emph{Applied Sciences}, 2022.

\bibitem[Han et~al.(2024)Han, Kim, Kang, Kim, Lee, and Lee]{sim_vul_sm}
Sungmin Han, Miju Kim, Jaesik Kang, Kwangsoo Kim, Seungwoon Lee, and Sangkyun Lee.
\newblock Similarity-based source code vulnerability detection leveraging transformer architecture: Harnessing cross- attention for hierarchical analysis.
\newblock \emph{IEEE Access}, 2024.

\bibitem[Jain and Wallace(2019)]{attention_is_not_explain}
Sarthak Jain and Byron~C. Wallace.
\newblock Attention is not explanation.
\newblock In \emph{NAACL-HLT}, 2019.

\bibitem[Jawahar et~al.(2019)Jawahar, Sagot, and Seddah]{level_of_semantics3}
Ganesh Jawahar, Beno{\^\i}t Sagot, and Djam{\'e} Seddah.
\newblock What does {BERT} learn about the structure of language?
\newblock In \emph{ACL}, 2019.

\bibitem[Kendall(1955)]{kendall1948rank}
Maurice~George Kendall.
\newblock \emph{Rank correlation methods.}
\newblock C. Griffin, 1955.

\bibitem[Lee and Han(2022)]{LibraCAM}
Sangkyun Lee and Sungmin Han.
\newblock {Libra-CAM:} an activation-based attribution based on the linear approximation of deep neural nets and threshold calibration.
\newblock In \emph{IJCAI}, 2022.

\bibitem[Liu et~al.(2019)Liu, Ott, Goyal, Du, Joshi, Chen, Levy, Lewis, Zettlemoyer, and Stoyanov]{roberta}
Yinhan Liu, Myle Ott, Naman Goyal, Jingfei Du, Mandar Joshi, Danqi Chen, Omer Levy, Mike Lewis, Luke Zettlemoyer, and Veselin Stoyanov.
\newblock {RoBERTa}: A robustly optimized {BERT} pretraining approach.
\newblock \emph{arXiv:1907.11692}, 2019.

\bibitem[Maas et~al.(2011)Maas, Daly, Pham, Huang, Ng, and Potts]{imdb}
Andrew~L. Maas, Raymond~E. Daly, Peter~T. Pham, Dan Huang, Andrew~Y. Ng, and Christopher Potts.
\newblock Learning word vectors for sentiment analysis.
\newblock In \emph{ACL-HLT}, 2011.

\bibitem[Martins and Astudillo(2016)]{attn_xai_2}
Andr\'{e} F.~T. Martins and Ram\'{o}n~F. Astudillo.
\newblock From softmax to sparsemax: A sparse model of attention and multi-label classification.
\newblock In \emph{ICML}, 2016.

\bibitem[Modarressi et~al.(2022)Modarressi, Fayyaz, Yaghoobzadeh, and Pilehvar]{globenc}
Ali Modarressi, Mohsen Fayyaz, Yadollah Yaghoobzadeh, and Mohammad~Taher Pilehvar.
\newblock {G}lob{E}nc: Quantifying global token attribution by incorporating the whole encoder layer in transformers.
\newblock In \emph{NAACL}, 2022.

\bibitem[Mohebbi et~al.(2023)Mohebbi, Zuidema, Chrupa{\l}a, and Alishahi]{rollout2}
Hosein Mohebbi, Willem Zuidema, Grzegorz Chrupa{\l}a, and Afra Alishahi.
\newblock Quantifying context mixing in transformers.
\newblock In \emph{EACL}, 2023.

\bibitem[Montavon et~al.(2019)Montavon, Binder, Lapuschkin, Samek, and M{\"u}ller]{lrp_rule}
Gr{\'e}goire Montavon, Alexander Binder, Sebastian Lapuschkin, Wojciech Samek, and Klaus-Robert M{\"u}ller.
\newblock \emph{Layer-wise relevance propagation: an overview}, pages 193--209.
\newblock Springer International Publishing, 2019.

\bibitem[Nguyen(2018)]{aopc1}
Dong Nguyen.
\newblock Comparing automatic and human evaluation of local explanations for text classification.
\newblock In \emph{NAACL-HLT}, 2018.

\bibitem[Pascual et~al.(2021)Pascual, Brunner, and Wattenhofer]{emnlp_review_telling_bert}
Damian Pascual, Gino Brunner, and Roger Wattenhofer.
\newblock Telling {BERT}{'}s full story: from local attention to global aggregation.
\newblock In \emph{EACL}, 2021.

\bibitem[Petsiuk(2018)]{deletion}
V~Petsiuk.
\newblock {Rise:} randomized input sampling for explanation of black-box models.
\newblock In \emph{BMVC}, 2018.

\bibitem[Qiang et~al.(2022)Qiang, Pan, Li, Li, Jang, and Zhu]{Attcat}
Yao Qiang, Deng Pan, Chengyin Li, Xin Li, Rhongho Jang, and Dongxiao Zhu.
\newblock {AttCAT:} explaining transformers via attentive class activation tokens.
\newblock In \emph{NIPS}, 2022.

\bibitem[Radford et~al.(2019)Radford, Wu, Child, Luan, Amodei, Sutskever, et~al.]{gpt2}
Alec Radford, Jeffrey Wu, Rewon Child, David Luan, Dario Amodei, Ilya Sutskever, et~al.
\newblock Language models are unsupervised multitask learners.
\newblock OpenAI blog, 2019.
\newblock URL \url{https://cdn.openai.com/better-language-models/language_models_are_unsupervised_multitask_learners.pdf}.

\bibitem[Rong et~al.(2022)Rong, Leemann, Borisov, Kasneci, and Kasneci]{morf_lerf_consistency}
Yao Rong, Tobias Leemann, Vadim Borisov, Gjergji Kasneci, and Enkelejda Kasneci.
\newblock A consistent and efficient evaluation strategy for attribution methods.
\newblock In \emph{ICML}, 2022.

\bibitem[Sanh et~al.(2019)Sanh, Debut, Chaumond, and Wolf]{distlbert}
Victor Sanh, Lysandre Debut, Julien Chaumond, and Thomas Wolf.
\newblock {DistilBERT}, a distilled version of {BERT}: smaller, faster, cheaper and lighter.
\newblock \emph{arXiv:1910.01108}, 2019.

\bibitem[Selvaraju et~al.(2017)Selvaraju, Cogswell, Das, Vedantam, Parikh, and Batra]{GradCAM}
Ramprasaath~R. Selvaraju, Michael Cogswell, Abhishek Das, Ramakrishna Vedantam, Devi Parikh, and Dhruv Batra.
\newblock {Grad-CAM:} visual explanations from deep networks via gradient-based localization.
\newblock In \emph{ICCV}, 2017.

\bibitem[Shrikumar et~al.(2017)Shrikumar, Greenside, and Kundaje]{lodss}
Avanti Shrikumar, Peyton Greenside, and Anshul Kundaje.
\newblock Learning important features through propagating activation differences.
\newblock In \emph{ICML}, 2017.

\bibitem[Socher et~al.(2013)Socher, Perelygin, Wu, Chuang, Manning, Ng, and Potts]{sst}
Richard Socher, Alex Perelygin, Jean Wu, Jason Chuang, Christopher~D. Manning, Andrew Ng, and Christopher Potts.
\newblock Recursive deep models for semantic compositionality over a sentiment treebank.
\newblock In \emph{EMNLP}, 2013.

\bibitem[{The White House}(2023)]{biden}
{The White House}.
\newblock Executive order on the safe, secure, and trustworthy development and use of artificial intelligence.
\newblock The White House Presidential Actions, 2023.

\bibitem[Touvron et~al.(2023)Touvron, Martin, Stone, Albert, Almahairi, Babaei, Bashlykov, Batra, Bhargava, Bhosale, et~al.]{llama2}
Hugo Touvron, Louis Martin, Kevin Stone, Peter Albert, Amjad Almahairi, Yasmine Babaei, Nikolay Bashlykov, Soumya Batra, Prajjwal Bhargava, Shruti Bhosale, et~al.
\newblock Llama 2: Open foundation and fine-tuned chat models.
\newblock \emph{arXiv preprint arXiv:2307.09288}, 2023.

\bibitem[Turton et~al.(2021)Turton, Smith, and Vinson]{level_of_semantics1}
Jacob Turton, Robert~Elliott Smith, and David Vinson.
\newblock Deriving contextualised semantic features from bert (and other transformer model) embeddings.
\newblock In \emph{Proceedings of the 6th Workshop on Representation Learning for NLP (RepL4NLP-2021)}, 2021.

\bibitem[Vaswani et~al.(2017)Vaswani, Shazeer, Parmar, Uszkoreit, Jones, Gomez, Kaiser, and Polosukhin]{transformer}
Ashish Vaswani, Noam Shazeer, Niki Parmar, Jakob Uszkoreit, Llion Jones, Aidan~N Gomez, \L~ukasz Kaiser, and Illia Polosukhin.
\newblock Attention is all you need.
\newblock In \emph{NIPS}, 2017.

\bibitem[Voita et~al.(2019)Voita, Talbot, Moiseev, Sennrich, and Titov]{PartialLRP}
Elena Voita, David Talbot, Fedor Moiseev, Rico Sennrich, and Ivan Titov.
\newblock Analyzing multi-head self-attention: Specialized heads do the heavy lifting, the rest can be pruned.
\newblock In \emph{ACL}, 2019.

\bibitem[Wang et~al.(2020)Wang, Wang, Du, Yang, Zhang, Ding, Mardziel, and Hu]{ScoreCAM}
Haofan Wang, Zifan Wang, Mengnan Du, Fan Yang, Zijian Zhang, Sirui Ding, Piotr Mardziel, and Xia Hu.
\newblock {Score-CAM:} score-weighted visual explanations for convolutional neural networks.
\newblock In \emph{CVPR (workshop)}, 2020.

\bibitem[Wiegreffe and Pinter(2019)]{attention_not_not_explain}
Sarah Wiegreffe and Yuval Pinter.
\newblock Attention is not not explanation.
\newblock In \emph{EMNLP-IJCNLP}, 2019.

\bibitem[Wolf et~al.(2020)Wolf, Debut, Sanh, Chaumond, Delangue, Moi, Cistac, Rault, Louf, Funtowicz, Davison, Shleifer, von Platen, Ma, Jernite, Plu, Xu, Le~Scao, Gugger, Drame, Lhoest, and Rush]{wolf2019huggingface}
Thomas Wolf, Lysandre Debut, Victor Sanh, Julien Chaumond, Clement Delangue, Anthony Moi, Pierric Cistac, Tim Rault, Remi Louf, Morgan Funtowicz, Joe Davison, Sam Shleifer, Patrick von Platen, Clara Ma, Yacine Jernite, Julien Plu, Canwen Xu, Teven Le~Scao, Sylvain Gugger, Mariama Drame, Quentin Lhoest, and Alexander Rush.
\newblock Transformers: State-of-the-art natural language processing.
\newblock In \emph{EMNLP}, 2020.

\bibitem[Zhang et~al.(2015)Zhang, Zhao, and LeCun]{amazon_yelp}
Xiang Zhang, Junbo Zhao, and Yann LeCun.
\newblock Character-level convolutional networks for text classification.
\newblock In \emph{NIPS}, 2015.

\end{thebibliography}

\newpage
\onecolumn

\appendix
\section*{\Large Appendix}

Most experiments were conducted on a system equipped with an Intel Xeon Silver 4214 CPU, 32GB of RAM, and an NVIDIA GeForce RTX 2080Ti GPU, with CUDA v10.2 used for GPU acceleration. To handle large-scale models such as Llama-2~\citep{llama2}, we employed an NVIDIA A100 GPU.

Baseline methods were implemented using the open-source code provided by \citet{Attcat} and \citet{TIS}.
We evaluated our approach on five publicly available NLP datasets for text classification: Amazon Polarity~\citep{amazon_yelp}, Yelp Polarity~\citep{amazon_yelp}, SST-2~\citep{sst}, IMDB~\citep{imdb}, and AG News~\citep{agnews}.

\section{Transformer Models}\label{appnedix.models}

\begin{table}[!ht]
\centering
\caption{Test accuracy of transformer-based text classification models used in our experiments.}\label{tbl:model_performance}
%\small
\newcolumntype{C}[1]{>{\centering\arraybackslash}m{#1}}
\begin{tabular}{c|c|c|c|c|c}\toprule
 \multirow{2}{*}{Model}& \multicolumn{5}{c}{Dataset}\\\cline{2-6}
 & \multicolumn{1}{c|}{Amazon} & \multicolumn{1}{c|}{Yelp} & \multicolumn{1}{c|}{SST2}& \multicolumn{1}{c|}{IMDB} & \multicolumn{1}{c}{AgNews} \\\midrule%\cline{2-7}

BERT$_{\text{base}}$ & 0.946 & 0.956 & 0.924 & 0.930 & 0.941 \\
DistilBERT & 0.945 & 0.962 & 0.891 & 0.928 & 0.947 \\
RoBERTa & 0.953 & 0.982 & 0.940 & 0.953 & 0.947 \\
GPT-2 & 0.968 & 0.977 & 0.921 & 0.877 & 0.949 \\
Llama-2 & 0.975 & 0.981 & 0.958 & 0.967 & 0.945 \\\bottomrule
\end{tabular}
\end{table}

We conducted our experiments using five types of transformer-based models: BERT$_{\text{base}}$~\citep{bert}, DistilBERT~\citep{distlbert}, RoBERTa~\citep{roberta}, GPT-2~\citep{gpt2}, and Llama-2 (7B)~\citep{llama2}.
We used pre-trained versions of these models from Hugging Face~\citep{wolf2019huggingface}. 
Table~\ref{tbl:model_performance} presents the accuracies of each pre-trained model on the five datasets used in our experiments.

\section{Baseline Methods}\label{appnedix.baseline}

We summarize below the baseline methods used in our experiments, along with brief descriptions:
\begin{itemize}
\item RawAtt (Raw Attention): Uses the raw attention weights from the final transformer layer as token importance scores.
\item Rollout~\citep{Rollout}: Aggregates attention across layers by recursively multiplying attention matrices, capturing how information flows through the entire network.
\item Att-grads~\citep{Gradsam}: Computes the gradient of the output with respect to attention weights, reflecting how changes in attention affect the model prediction.
\item Att $\times$ Att-grads~\citep{Gradsam}: Multiplies attention scores with their gradients to highlight tokens that are both attended to and influential to the output.
\item Grad-SAM~\citep{Gradsam}: A refinement of attention-gradient methods, applying self-attention map gradients in a Grad-CAM-style formulation [8] for better localization of influential tokens.
\item Full LRP~\citep{FLRP}: Applies Layer-wise Relevance Propagation (LRP) across the entire transformer model, decomposing predictions into token-wise relevance scores based on conservation principles.
\item Partial LRP~\citep{PartialLRP}: Restricts LRP computation to the final classification layers, assuming most interpretability-relevant information is captured in later stages.
\item TransAtt~\citep{Transatt}: Proposes a transformer-specific LRP mechanism that refines attribution by accounting for residual connections and non-linearities across layers.
\item CAT~\citep{Attcat}: Inspired by Grad-CAM~\citep{GradCAM}, CAT computes token-level attributions by combining gradients and encoded feature activations, producing class-specific importance scores that reflect individual token contributions.
\item AttCAT~\citep{Attcat}:  Extends CAT by integrating self-attention weights to account for inter-token interactions, enhancing the faithfulness of token-level explanations in Transformers.
\item TIS~\citep{TIS}: Computes token-level saliency by measuring the change in model output—similar to Score-CAM~\citep{ScoreCAM}—when subsets of input tokens are masked. These subsets are selected based on centroids obtained from K-means clustering of activations, while special tokens (e.g., the classification [CLS] token) are preserved.
\end{itemize}

\section{Faithfulness of Attribution}\label{appnedix.faithful}

\paragraph{Additional Experimental Results for the BERT$_{\text{base}}$ Model}

\begin{figure*}[!ht]
\centering
\includegraphics[width=0.95\textwidth]{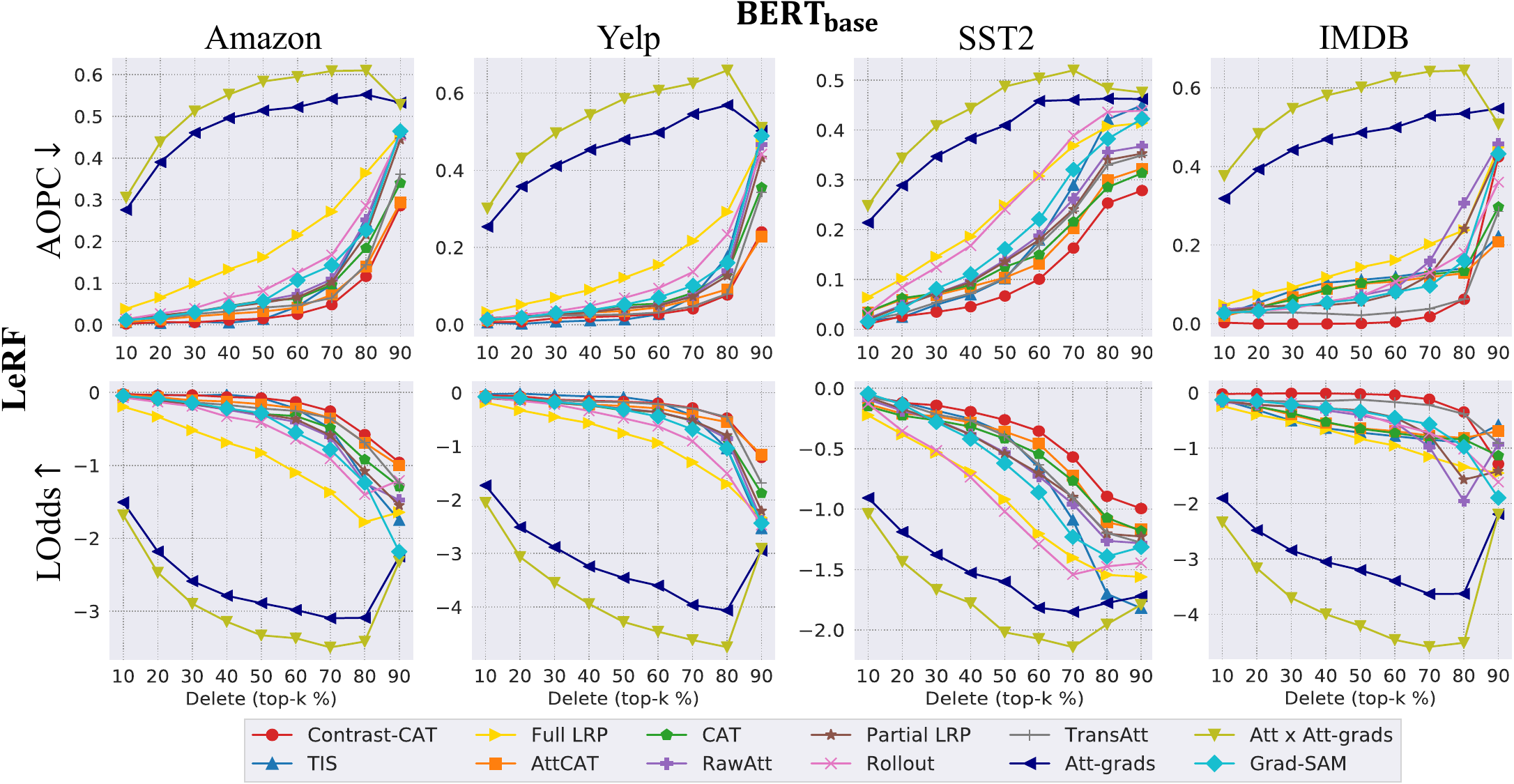}
\caption{Quantitative comparison of the faithfulness evaluation of \mymethod and other attribution methods, measured under the LeRF (Least Relevant First) setting.}
\label{fig:plot.bert_quantitative_lerf}
\end{figure*}

Figure~\ref{fig:plot.bert_quantitative_lerf} shows the faithfulness evaluation results under the LeRF setting, corresponding to the results labeled as (B) LeRF in Table~\ref{tbl:all.measures} of our main manuscript. Table~\ref{tbl:all.measures.bert.agnews} presents the faithfulness evaluation results of attribution methods on the AgNews dataset using the BERT$_{\text{base}}$ model.

As shown in Table~\ref{tbl:all.measures.bert.agnews}, \mymethod demonstrates consistently superior attribution quality on the AgNews dataset compared to other competing methods, similar to the results in Table~\ref{tbl:all.measures}.

\begin{table*}[tb]
\centering
\caption{AUC values for the faithfulness evaluation of attribution methods using the \textbf{BERT$_{\text{base}}$} model on the AgNews dataset under the MoRF (Most Relevant First) and LeRF (Least Relevant First) settings. The best and the second-best cases are in boldface and underlined, respectively.}\label{tbl:all.measures.bert.agnews}
%\small
\newcolumntype{C}[1]{>{\centering\arraybackslash}m{#1}}
\begin{tabular}{c|c|c|c|c}\toprule
\multirow{2}{*}{Setting} & \multicolumn{2}{c|}{MoRF} & \multicolumn{2}{c}{LeRF} \\
& \multicolumn{2}{c|}{(Most Relevant First)} & \multicolumn{2}{c}{(Least Relevant First)} \\\midrule
Method & AOPC$\uparrow$ & LOdds$\downarrow$ & AOPC$\downarrow$ & LOdds$\uparrow$\\
\midrule

RawAtt & 0.268 & 0.580 & 0.152 & 0.663 \\ 
Rollout & 0.300 & 0.532 & 0.184 & 0.639 \\ 
Att-grads & 0.099 & 0.728 & 0.331 & 0.461 \\ 
Att$\times$Att-grads & 0.084 & 0.739 & 0.379 & 0.394 \\ 
Grad-SAM & 0.270 & 0.578 & 0.180 & 0.632 \\ 
Full LRP & 0.234 & 0.604 & 0.199 & 0.623 \\ 
Partial LRP & 0.294 & 0.555 & 0.135 & 0.681 \\ 
TransAtt & 0.347 & 0.499 & \underline{0.105} & \underline{0.714} \\ 
CAT & 0.273 & 0.556 & 0.137 & 0.680 \\ 
AttCAT & 0.289 & 0.536 & 0.126 & 0.692 \\ 
TIS & \underline{0.354} & \underline{0.473} & 0.143 & 0.674 \\ 
\mymethod & \textbf{0.434} & \textbf{0.363} & \textbf{0.093} & \textbf{0.723} \\\bottomrule

\end{tabular}
\end{table*}

\paragraph{Experimental Results for Other Models}

We conducted the faithfulness evaluation of attribution methods, as detailed in Section~\ref{exp.faithful}, using the DistilBERT~\citep{distlbert}, RoBERTa~\citep{roberta}, GPT-2~\citep{gpt2}, and Llama-2~\citep{llama2} models.
In these experiments, we compared \mymethod against five different attribution methods: RawAtt and Rollout (attention-based methods), and CAT, AttCAT, and TIS (activation-based methods).

Table~\ref{tbl:all.measures.distil} reports results for DistilBERT, Table~\ref{tbl:all.measures.roberta} for the RoBERTa model, Table~\ref{tbl:all.measures.gpt2} for GPT-2, and Table~\ref{tbl:all.measures.llama2} Llama-2.
Additionally, Table~\ref{tbl:all.measures.allmodels.agnews} presents results on the AgNews dataset across all four models.
The results demonstrate the superior attribution quality of \mymethod across different datasets and models. 

Specifically, for the DistilBERT model, average improvements across different datasets are $\times 1.31$ in AOPC and $\times 2.39$ in LOdds compared to the second-best methods under the MoRF setting. Under the LeRF setting, \mymethod shows average improvements in AUC values for AOPC and LOdds by $\times 1.39$ and $\times 1.07$, respectively.

For the RoBERTa model, the average improvements are $\times 1.61$ in AOPC and $\times 2.97$ in LOdds under the MoRF setting, with AUC improvements of $\times 2.07$ and $\times 1.12$ in AOPC and LOdds, respectively, under the LeRF setting.
Similarly, for the GPT-2 model, the average improvements across datasets are $\times 2.78$ in AOPC and $\times 3.37$ in LOdds under the MoRF setting. For the LeRF setting, \mymethod demonstrates average improvements of $\times 3.80$ in AOPC and $\times 1.39$ in LOdds.

For the Llama-2 model, which is a large-scale transformer-based model, \mymethod outperforms competing attribution methods across different datasets. Under the MoRF setting, it achieves average improvements of $\times 1.74$ in AOPC and $\times 1.66$ in LOdds compared to the second-best method. In the LeRF setting, \mymethod shows improvements of $\times 1.10$ in AOPC and $\times 1.06$ in LOdds.

These results are consistent with those presented in Figure~\ref{fig:plot.quantitative} and Table~\ref{tbl:all.measures} in our main manuscript, further confirming the superiority of \mymethod in generating faithful attribution maps while highlighting the robustness of Contrast-CAT across various architectures and datasets.

\begin{table*}[!ht]
\centering
\caption{AUC values of the faithfulness evaluation conducted on the \textbf{DistilBERT} model. The best and the second-best cases are in boldface and underlined, respectively.}\label{tbl:all.measures.distil}
%\small
\newcolumntype{C}[1]{>{\centering\arraybackslash}m{#1}}
\begin{tabular}{c|c|c|c|c|c|c|c|c}\toprule

\multicolumn{9}{c}{(A) MoRF (Most Relevant First)}\\\bottomrule
 Dataset& \multicolumn{2}{c|}{Amazon} & \multicolumn{2}{c|}{Yelp} & \multicolumn{2}{c|}{SST2}& \multicolumn{2}{c}{IMDB} \\\midrule%\cline{2-7}
 Method & AOPC$\uparrow$ & LOdds$\downarrow$ & AOPC$\uparrow$ & LOdds$\downarrow$ & AOPC$\uparrow$ & LOdds$\downarrow$ & AOPC$\uparrow$ & LOdds$\downarrow$ \\\midrule
 
RawAtt & 0.360 & 0.557 & 0.306 & 0.618 & 0.363 & 0.531 & 0.172 & 0.729 \\ 
Rollout & 0.307 & 0.638 & 0.242 & 0.676 & 0.322 & 0.587 & 0.231 & 0.700 \\
CAT & 0.521 & 0.361 & 0.528 & 0.334 & 0.469 & 0.392 & 0.625 & 0.235 \\
AttCAT & \underline{0.532} & \underline{0.341} & \underline{0.570} & \underline{0.278} & \underline{0.480} & \underline{0.376} & \underline{0.638} & \underline{0.217} \\
TIS & 0.436 & 0.448 & 0.406 & 0.476 & 0.394 & 0.467 & 0.428 & 0.487 \\
\mymethod & \textbf{0.720} & \textbf{0.108} & \textbf{0.727} & \textbf{0.106} & \textbf{0.685} & \textbf{0.137} & \textbf{0.752} & \textbf{0.101} \\\cline{1-9}
\toprule
\multicolumn{9}{c}{(B) LeRF (Least Relevant First)}\\\bottomrule
Dataset& \multicolumn{2}{c|}{Amazon} & \multicolumn{2}{c|}{Yelp} & \multicolumn{2}{c|}{SST2}& \multicolumn{2}{c}{IMDB} \\\midrule%\cline{2-7}
 Method & AOPC$\downarrow$ & LOdds$\uparrow$ & AOPC$\downarrow$ & LOdds$\uparrow$ & AOPC$\downarrow$ & LOdds$\uparrow$ & AOPC$\downarrow$ & LOdds$\uparrow$ \\\midrule

RawAtt & 0.174 & 0.626 & 0.122 & 0.649 & 0.283 & 0.508 & 0.121 & 0.673 \\ 
Rollout & 0.181 & 0.606 & 0.112 & 0.655 & 0.328 & 0.429 & 0.090 & 0.706 \\ 
CAT & 0.119 & 0.678 & 0.065 & 0.708 & 0.248 & 0.536 & 0.028 & 0.773 \\ 
AttCAT & \underline{0.098} & \underline{0.703} & \underline{0.028} & \underline{0.764} & \underline{0.234} & \underline{0.549} & \underline{0.016} & \underline{0.787} \\ 
TIS & 0.162 & 0.637 & 0.113 & 0.669 & 0.315 & 0.478 & 0.089 & 0.708 \\ 
\mymethod & \textbf{0.068} & \textbf{0.737} & \textbf{0.020} & \textbf{0.779} & \textbf{0.142} & \textbf{0.669} & \textbf{0.015} & \textbf{0.788} \\\bottomrule
\end{tabular}
\end{table*}

\begin{table*}[!ht]
\centering
\caption{AUC values of the faithfulness evaluation conducted on the \textbf{RoBERTa} model. The best and the second-best cases are in boldface and underlined, respectively.}\label{tbl:all.measures.roberta}
%\small
\newcolumntype{C}[1]{>{\centering\arraybackslash}m{#1}}
\begin{tabular}{c|c|c|c|c|c|c|c|c}\toprule

\multicolumn{9}{c}{(A) MoRF (Most Relevant First)}\\\bottomrule
 Dataset& \multicolumn{2}{c|}{Amazon} & \multicolumn{2}{c|}{Yelp} & \multicolumn{2}{c|}{SST2}& \multicolumn{2}{c}{IMDB} \\\midrule%\cline{2-7}
 Method & AOPC$\uparrow$ & LOdds$\downarrow$ & AOPC$\uparrow$ & LOdds$\downarrow$ & AOPC$\uparrow$ & LOdds$\downarrow$ & AOPC$\uparrow$ & LOdds$\downarrow$ \\\midrule
 
RawAtt & 0.245 & 0.615 & 0.164 & 0.713 & 0.260 & 0.619 & 0.272 & 0.676 \\ 
Rollout & 0.188 & 0.660 & 0.153 & 0.717 & 0.195 & 0.653 & 0.274 & 0.657 \\
CAT & 0.287 & 0.557 & 0.357 & 0.526 & 0.461 & 0.410 & 0.452 & 0.464 \\
AttCAT & 0.274 & 0.568 & 0.347 & 0.532 & 0.454 & 0.416 & 0.449 & 0.467 \\
TIS & \underline{0.354} & \underline{0.503} & \underline{0.394} & \underline{0.503} & \underline{0.524} & \underline{0.372} & \underline{0.520} & \underline{0.411} \\
\mymethod & \textbf{0.688} & \textbf{0.140} & \textbf{0.684} & \textbf{0.160} & \textbf{0.686} & \textbf{0.160} & \textbf{0.738} & \textbf{0.131} \\\cline{1-9}
\toprule
\multicolumn{9}{c}{(B) LeRF (Least Relevant First)}\\\bottomrule
Dataset& \multicolumn{2}{c|}{Amazon} & \multicolumn{2}{c|}{Yelp} & \multicolumn{2}{c|}{SST2}& \multicolumn{2}{c}{IMDB} \\\midrule%\cline{2-7}
 Method & AOPC$\downarrow$ & LOdds$\uparrow$ & AOPC$\downarrow$ & LOdds$\uparrow$ & AOPC$\downarrow$ & LOdds$\uparrow$ & AOPC$\downarrow$ & LOdds$\uparrow$ \\\midrule

RawAtt & 0.218 & 0.586 & 0.177 & 0.581 & 0.323 & 0.457 & 0.157 & 0.556 \\ 
Rollout & 0.303 & 0.514 & 0.197 & 0.569 & 0.443 & 0.311 & 0.184 & 0.531 \\ 
CAT & 0.200 & 0.606 & 0.127 & 0.674 & 0.141 & 0.676 & 0.077 & 0.704 \\ 
AttCAT & 0.200 & 0.604 & 0.124 & \underline{0.677} & \underline{0.137} & 0.678 & 0.077 & 0.709 \\ 
TIS & \underline{0.191} & \underline{0.613} & \underline{0.119} & 0.669 & 0.143 & \underline{0.679} & \underline{0.076} & \underline{0.712} \\ 
\mymethod & \textbf{0.065} & \textbf{0.741} & \textbf{0.052} & \textbf{0.771} & \textbf{0.085} & \textbf{0.738} & \textbf{0.053} & \textbf{0.749} \\\bottomrule
\end{tabular}
\end{table*}

\begin{table*}[!ht]
\centering
\caption{AUC values of the faithfulness evaluation conducted on the \textbf{GPT-2} model. The best and the second-best cases are in boldface and underlined, respectively. N/A indicates that the method is not applicable to GPT-2.}\label{tbl:all.measures.gpt2}
%\small
\newcolumntype{C}[1]{>{\centering\arraybackslash}m{#1}}
\begin{tabular}{c|c|c|c|c|c|c|c|c}\toprule

\multicolumn{9}{c}{(A) MoRF (Most Relevant First)}\\\bottomrule
 Dataset& \multicolumn{2}{c|}{Amazon} & \multicolumn{2}{c|}{Yelp} & \multicolumn{2}{c|}{SST2}& \multicolumn{2}{c}{IMDB} \\\midrule%\cline{2-7}
 Method & AOPC$\uparrow$ & LOdds$\downarrow$ & AOPC$\uparrow$ & LOdds$\downarrow$ & AOPC$\uparrow$ & LOdds$\downarrow$ & AOPC$\uparrow$ & LOdds$\downarrow$ \\\midrule
 
RawAtt & 0.385 & 0.622 & 0.138 & 0.690 & \underline{0.303} & \underline{0.420} & \underline{0.163} & \underline{0.699} \\ 
Rollout & 0.320 & 0.684 & 0.138 & 0.690 & \underline{0.303} & \underline{0.420} & \underline{0.163} & \underline{0.699} \\
CAT & 0.505 & 0.392 & 0.177 & 0.653 & 0.243 & 0.617 & 0.042 & 0.775 \\
AttCAT & \underline{0.541} & \underline{0.345} & \underline{0.186} & \underline{0.647} & 0.221 & 0.662 & 0.043 & 0.775 \\ 
TIS & N/A & N/A & N/A & N/A & N/A & N/A & N/A & N/A \\
\mymethod & \textbf{0.744} & \textbf{0.136} & \textbf{0.617} & \textbf{0.188} & \textbf{0.636} & \textbf{0.188} & \textbf{0.706} & \textbf{0.132} \\\cline{1-9}
\toprule
\multicolumn{9}{c}{(B) LeRF (Least Relevant First)}\\\bottomrule
Dataset& \multicolumn{2}{c|}{Amazon} & \multicolumn{2}{c|}{Yelp} & \multicolumn{2}{c|}{SST2}& \multicolumn{2}{c}{IMDB} \\\midrule%\cline{2-7}
 Method & AOPC$\downarrow$ & LOdds$\uparrow$ & AOPC$\downarrow$ & LOdds$\uparrow$ & AOPC$\downarrow$ & LOdds$\uparrow$ & AOPC$\downarrow$ & LOdds$\uparrow$ \\\midrule

RawAtt & 0.193 & 0.513 & \underline{0.200} & \underline{0.524} & \underline{0.391} & 0.350 & \underline{0.200} & \underline{0.679} \\ 
Rollout & 0.215 & 0.472 & \underline{0.200} & \underline{0.524} & \underline{0.391} & 0.350 & \underline{0.200} & \underline{0.679} \\ 
CAT & 0.164 & 0.584 & 0.247 & 0.434 & 0.492 & 0.321 & 0.703 & 0.199 \\ 
AttCAT & \underline{0.129} & \underline{0.646} & 0.216 & 0.488 & 0.506 & \underline{0.359} & 0.679 & 0.231 \\ 
TIS & N/A & N/A & N/A & N/A & N/A & N/A & N/A & N/A \\
\mymethod & \textbf{0.093} & \textbf{0.696} & \textbf{0.062} & \textbf{0.731} & \textbf{0.206} & \textbf{0.700} & \textbf{0.023} & \textbf{0.790} \\ \bottomrule
\end{tabular}
\end{table*}

\begin{table*}[!ht]
\centering
\caption{AUC values of the faithfulness evaluation conducted on the \textbf{Llama-2} model. The best and the second-best cases are in boldface and underlined, respectively. N/A indicates that the method is not applicable to Llama-2.}\label{tbl:all.measures.llama2}
%\small
\newcolumntype{C}[1]{>{\centering\arraybackslash}m{#1}}
\begin{tabular}{c|c|c|c|c|c|c|c|c}\toprule

\multicolumn{9}{c}{(A) MoRF (Most Relevant First)}\\\bottomrule
 Dataset& \multicolumn{2}{c|}{Amazon} & \multicolumn{2}{c|}{Yelp} & \multicolumn{2}{c|}{SST2}& \multicolumn{2}{c}{IMDB} \\\midrule%\cline{2-7}
 Method & AOPC$\uparrow$ & LOdds$\downarrow$ & AOPC$\uparrow$ & LOdds$\downarrow$ & AOPC$\uparrow$ & LOdds$\downarrow$ & AOPC$\uparrow$ & LOdds$\downarrow$ \\\midrule
 
RawAtt & 0.174 & 0.622 & 0.095 & 0.712 & 0.244 & 0.561 & 0.132 & 0.731 \\ 
Rollout & 0.174 & 0.622 & 0.095 & 0.712 & 0.244 & 0.561 & 0.132 & 0.731 \\ 
CAT & \underline{0.355} & \underline{0.447} & 0.315 & 0.551 & 0.389 & \underline{0.464} & 0.363 & 0.631 \\ 
AttCAT & 0.354 & \underline{0.447} & \underline{0.339} & \underline{0.527} & \underline{0.395} & 0.466 & \underline{0.368} & \underline{0.621} \\ 
TIS & N/A & N/A & N/A & N/A & N/A & N/A & N/A & N/A \\
\mymethod & \textbf{0.554} & \textbf{0.228} & \textbf{0.727} & \textbf{0.141} & \textbf{0.622} & \textbf{0.202} & \textbf{0.620} & \textbf{0.514} \\\cline{1-9}
\toprule
\multicolumn{9}{c}{(B) LeRF (Least Relevant First)}\\\bottomrule
Dataset& \multicolumn{2}{c|}{Amazon} & \multicolumn{2}{c|}{Yelp} & \multicolumn{2}{c|}{SST2}& \multicolumn{2}{c}{IMDB} \\\midrule%\cline{2-7}
 Method & AOPC$\downarrow$ & LOdds$\uparrow$ & AOPC$\downarrow$ & LOdds$\uparrow$ & AOPC$\downarrow$ & LOdds$\uparrow$ & AOPC$\downarrow$ & LOdds$\uparrow$ \\\midrule

RawAtt & \textbf{0.217} & 0.558 & \underline{0.177} & \underline{0.658} & 0.334 & 0.384 & 0.203 & 0.717 \\  
Rollout & \textbf{0.217} & 0.558 & \underline{0.177} & \underline{0.658} & 0.334 & 0.384 & 0.203 & 0.717 \\ 
CAT & 0.241 & 0.557 & 0.497 & 0.355 & \underline{0.305} & \underline{0.488} & 0.155 & 0.725 \\ 
AttCAT & \underline{0.233} & \underline{0.567} & 0.463 & 0.400 & 0.313 & 0.475 & \underline{0.150} & \underline{0.732} \\ 
TIS & N/A & N/A & N/A & N/A & N/A & N/A & N/A & N/A \\
\mymethod & \underline{0.233} & \textbf{0.577} & \textbf{0.170} & \textbf{0.672} & \textbf{0.218} & \textbf{0.583} & \textbf{0.146} & \textbf{0.735} \\ \bottomrule
\end{tabular}
\end{table*}

\begin{table*}[tb]
\centering
\caption{Faithfulness evaluation results of attribution methods conducted on the AgNews dataset using different models (DistilBERT, RoBERTa, GPT-2, and Llama-2) under MoRF (Most Relevant First) and LeRF (Least Relevant First) settings.}\label{tbl:all.measures.allmodels.agnews}
%\small
\newcolumntype{C}[1]{>{\centering\arraybackslash}m{#1}}
\begin{tabular}{c|c|c|c|c|c|c|c|c}\toprule
& \multicolumn{8}{c}{(A) MoRF (Most Relevant First)}\\\midrule
Model & \multicolumn{2}{c|}{DistilBERT} & \multicolumn{2}{c|}{RoBERTa} & \multicolumn{2}{c|}{GPT-2} & \multicolumn{2}{c}{Llama-2} \\\midrule
Method & AOPC$\uparrow$ & LOdds$\downarrow$ & AOPC$\uparrow$ & LOdds$\downarrow$ & AOPC$\uparrow$ & LOdds$\downarrow$ & AOPC$\uparrow$ & LOdds$\downarrow$ \\
\midrule

RawAtt & 0.218 & 0.669 & 0.352 & 0.566 & 0.174 & 0.554 & 0.163 & 0.656\\ 
Rollout & 0.316 & 0.620 & 0.181 & 0.673 & 0.174 & 0.554 & 0.163 & 0.656\\ 
CAT & 0.344 & 0.492 & 0.333 & 0.530 & 0.174 & 0.557 & 0.274 & 0.540\\ 
AttCAT & \underline{0.345} & \underline{0.487} & 0.330 & 0.540 & \underline{0.176} & \underline{0.575} & \underline{0.282} & \underline{0.533} \\ 
TIS & 0.323 & 0.556 & \underline{0.413} & \underline{0.456} & N/A & N/A & N/A & N/A \\ 
\mymethod & \textbf{0.452} & \textbf{0.382} & \textbf{0.680} & \textbf{0.169} & \textbf{0.350} & \textbf{0.256} & \textbf{0.473} & \textbf{0.284}\\ \midrule
& \multicolumn{8}{c}{(B) LeRF (Least Relevant First)}\\\midrule
Model & \multicolumn{2}{c|}{DistilBERT} & \multicolumn{2}{c|}{RoBERTa} & \multicolumn{2}{c|}{GPT-2} & \multicolumn{2}{c}{Llama-2} \\\midrule
Method & AOPC$\downarrow$ & LOdds$\uparrow$ & AOPC$\downarrow$ & LOdds$\uparrow$ & AOPC$\downarrow$ & LOdds$\uparrow$ & AOPC$\downarrow$ & LOdds$\uparrow$\\
\midrule

RawAtt & 0.225 & 0.609 & 0.188 & 0.580 & 0.178 & 0.484 & \underline{0.167} & \underline{0.609}\\ 
Rollout & 0.141 & 0.688 & 0.224 & 0.562 & 0.178 & 0.484 & \underline{0.167} & \underline{0.609}\\ 
CAT & \underline{0.072} & \textbf{0.752} & \underline{0.096} & \underline{0.699} & \underline{0.255} & \underline{0.409} & 0.178 & 0.599\\ 
AttCAT & \textbf{0.068} & \textbf{0.752} & 0.098 & 0.698 & 0.256 & 0.393 & 0.183 & 0.586 \\ 
TIS & 0.154 & 0.702 & 0.109 & 0.690 & N/A & N/A & N/A & N/A \\ 
\mymethod & \underline{0.072} & \underline{0.746} & \textbf{0.061} & \textbf{0.742} & \textbf{0.161} & \textbf{0.588} &\textbf{0.136} &\textbf{0.653} \\\bottomrule

\end{tabular}
\end{table*}

\clearpage
\section{Confidence of Attribution}\label{appnedix.confidence}

\begin{table}[!ht]
\centering
\caption{The results of confidence evaluation conducted on the \textbf{DistilBERT}, \textbf{RoBERTa}, \textbf{GPT-2}, and \textbf{Llama-2} models. Values below $0.05$ are marked in gray. N/A indicates that the method is not applicable to the given model.}\label{tbl:distil.all.k_tau}
%\small
\setlength{\tabcolsep}{8pt}
\begin{tabular}{c|c|c|c|c|c|c|c}\toprule
Model & Dataset & RawAtt & Rollout & CAT & AttCAT & TIS & \mymethod \\
\hline

\multirow{5}{*}{\rotatebox{90}{DistilBERT}} 
& Amazon & 1.00 & 1.00 & \cellcolor[gray]{0.8}$<$ 0.05 & \cellcolor[gray]{0.8}$<$ 0.05 & \cellcolor[gray]{0.8}$<$ 0.05 & \cellcolor[gray]{0.8}$<$ 0.05 \\ 
& Yelp & 1.00 & 1.00 & \cellcolor[gray]{0.8}$<$ 0.05 & \cellcolor[gray]{0.8}$<$ 0.05 & \cellcolor[gray]{0.8}$<$ 0.05 & \cellcolor[gray]{0.8}$<$ 0.05 \\ 
& SST2 & 1.00 & 1.00 & \cellcolor[gray]{0.8}$<$ 0.05 & \cellcolor[gray]{0.8}$<$ 0.05 & \cellcolor[gray]{0.8}$<$ 0.05 & \cellcolor[gray]{0.8}$<$ 0.05 \\ 
& IMDB & 1.00 & 1.00 & \cellcolor[gray]{0.8}$<$ 0.05 & \cellcolor[gray]{0.8}$<$ 0.05 & \cellcolor[gray]{0.8}$<$ 0.05 & \cellcolor[gray]{0.8}$<$ 0.05 \\ 
& AgNews & 1.00  & 1.00 & 0.069 & \cellcolor[gray]{0.8}$<$ 0.05 & \cellcolor[gray]{0.8}$<$ 0.05 & \cellcolor[gray]{0.8}$<$ 0.05 \\ \hline

\multirow{5}{*}{\rotatebox{90}{RoBERTa}} 
& Amazon & 1.00 & 1.00 & \cellcolor[gray]{0.8}$<$ 0.05 & \cellcolor[gray]{0.8}$<$ 0.05 & \cellcolor[gray]{0.8}$<$ 0.05 & \cellcolor[gray]{0.8}$<$ 0.05 \\ 
& Yelp & 1.00 & 1.00 & \cellcolor[gray]{0.8}$<$ 0.05 & \cellcolor[gray]{0.8}$<$ 0.05 & \cellcolor[gray]{0.8}$<$ 0.05 & \cellcolor[gray]{0.8}$<$ 0.05 \\ 
& SST2 & 1.00 & 1.00 & \cellcolor[gray]{0.8}$<$ 0.05 & \cellcolor[gray]{0.8}$<$ 0.05 & \cellcolor[gray]{0.8}$<$ 0.05 & \cellcolor[gray]{0.8}$<$ 0.05 \\ 
& IMDB & 1.00 & 1.00 & \cellcolor[gray]{0.8}$<$ 0.05 & \cellcolor[gray]{0.8}$<$ 0.05 & \cellcolor[gray]{0.8}$<$ 0.05 & \cellcolor[gray]{0.8}$<$ 0.05 \\ 
& AgNews & 1.00 & 1.00 & 0.050 & 0.054 & \cellcolor[gray]{0.8}$<$ 0.05 & \cellcolor[gray]{0.8}$<$ 0.05 \\ \hline

\multirow{5}{*}{\rotatebox{90}{GPT-2}} 
& Amazon & 1.00 & 1.00 & \cellcolor[gray]{0.8}$<$ 0.05 & \cellcolor[gray]{0.8}$<$ 0.05 & N/A & \cellcolor[gray]{0.8}$<$ 0.05 \\ 
& Yelp & 1.00 & 1.00 & \cellcolor[gray]{0.8}$<$ 0.05 & \cellcolor[gray]{0.8}$<$ 0.05 & N/A & \cellcolor[gray]{0.8}$<$ 0.05 \\ 
& SST2 & 1.00 & 1.00 & \cellcolor[gray]{0.8}$<$ 0.05 & \cellcolor[gray]{0.8}$<$ 0.05 & N/A & \cellcolor[gray]{0.8}$<$ 0.05 \\ 
& IMDB & 1.00 & 1.00 & \cellcolor[gray]{0.8}$<$ 0.05 & \cellcolor[gray]{0.8}$<$ 0.05 & N/A & \cellcolor[gray]{0.8}$<$ 0.05 \\ 
& AgNews & 1.00 & 1.00 & \cellcolor[gray]{0.8}$<$ 0.05 & 0.068 & N/A  & \cellcolor[gray]{0.8}$<$ 0.05 \\
\hline

\multirow{5}{*}{\rotatebox{90}{Llama-2}} 
& Amazon & 1.00 & 1.00 & \cellcolor[gray]{0.8}$<$ 0.05 & 0.104 & N/A & \cellcolor[gray]{0.8}$<$ 0.05 \\ 
& Yelp & 0.999 & 0.999 & \cellcolor[gray]{0.8}$<$ 0.05 & 0.091 & N/A & \cellcolor[gray]{0.8}$<$ 0.05 \\ 
& SST2 & 1.00 & 1.00 & \cellcolor[gray]{0.8}$<$ 0.05 & 0.090 & N/A & \cellcolor[gray]{0.8}$<$ 0.05 \\ 
& IMDB & 1.00 & 1.00 & \cellcolor[gray]{0.8}$<$ 0.05 & \cellcolor[gray]{0.8}$<$ 0.05 & N/A & \cellcolor[gray]{0.8}$<$ 0.05 \\ 
& AgNews & 1.00 & 1.00 & \cellcolor[gray]{0.8}$<$ 0.05 & 0.069 & N/A & 0.052 \\ 

\bottomrule
\end{tabular}
\end{table}

Table~\ref{tbl:distil.all.k_tau}  presents the confidence evaluation results for various attribution methods conducted on the DistilBERT, RoBERTa, GPT-2, and Llama-2 models. 
The results show that \mymethod consistently achieves average rank correlation values below $0.05$ across all datasets and models, except for the AgNews dataset. In the case of AgNews, although the value exceeded $0.05$, it remained very small at $0.052$.

\section{Ablation Study}\label{appnedix.abla}

\begin{table}[tb]
\centering
\caption{The table presents the average AUC values across multiple datasets (Amazon, Yelp, SST2, and IMDB), evaluated under the MoRF (Most Relevant First) setting using the BERT$_{\text{base}}$ model. These values reflect the impact of the components of \mymethod: multiple contrasting (M), attention score (A), and refinement via deletion test (R). The ``Improvement" columns indicate the performance gain compared to the baseline (without M, A, and R).}
\label{tbl:ablation}
\begin{tabular}{ccc|c|c|c|c}
\toprule
 \multicolumn{3}{c|}{Components} & \multirow{2}{*}{AOPC$\uparrow$} & \multirow{2}{*}{Improvement} & \multirow{2}{*}{LOdds$\downarrow$} & \multirow{2}{*}{Improvement}\\
 M & A & R & & & & \\\midrule
 \xmark & \xmark & \xmark & 0.456 & 1.00 & 0.375 & 1.00\\
 \cmark & \xmark & \xmark & 0.506 & $\times$1.11 & 0.323 & $\times$1.16\\
 \cmark & \cmark & \xmark & 0.556 & $\times$1.22 & 0.271 & $\times$1.38\\
 \cmark & \cmark & \cmark & 0.696 & $\times$1.53 & 0.127 & $\times$2.95\\
\bottomrule
\end{tabular}
\end{table}

Table~\ref{tbl:ablation} presents the average AUC values evaluated under the MoRF setting using the BERT$_{\text{base}}$ model for AOPC and LOdds across the four datasets (Amazon, Yelp, SST2, and IMDB) used in our main manuscript. The table illustrates the individual and combined impact of multiple contrasting (M), attention score (A), and refinement via deletion test (R) on the attribution quality of \mymethod.
For instance, applying multiple contrasting alone produces attribution maps with multiple references and generates an attribution map by simply averaging them, without incorporating the attention score or the refinement procedure.

The results demonstrate that each component contributes to overall performance, with the combined use of all three components yielding the most significant improvement.

\section{Efficiency of using a reference library}\label{appnedix.ref_lib_efficiency}

\begin{table*}[!ht]
\centering
\caption{Comparison of attribution quality between \mymethod with and without a reference library. The version using a reference library is labeled "Reference Library (Deterministic)", while the version without it is labeled "On-the-Fly (Constrained Random)". TIS is included as a baseline. The best results are highlighted in bold.}\label{tbl:perf_comparison_ref_lib}

\newcolumntype{C}[1]{>{\centering\arraybackslash}m{#1}}
\begin{tabular}{c|c|c}\toprule
 Method & AOPC$\uparrow$ & LOdds$\downarrow$ \\
\midrule
TIS (baseline) & 0.463 & 0.367 \\
On-the-Fly (Constrained Random) & 0.647, 0.009 (Mean, Std) & 0.159, 0.012 (Mean, Std)\\
Reference Library (Deterministic) & \textbf{0.654} & \textbf{0.157} \\\bottomrule
\end{tabular}
\end{table*}

\begin{table}[!ht]
\centering
\caption{Comparison of runtime between \mymethod with and without a reference library, denoted as "Reference Library" and "On-the-Fly", respectively. The average runtime per sample on the SST2 dataset is reported, with the best result shown in bold.}\label{tbl:time_comparison_ref_lib}
\begin{tabular}{c|c|c|c}\toprule
Method & TIS & On-the-Fly & Reference Library \\
\hline
Time & 22.669 & 4.782 & \textbf{2.134} \\\bottomrule
\end{tabular}
\end{table}

As described in Section~\ref{sec:multi-references}, Contrast-CAT utilizes a pre-built reference library to reduce its computational overhead associated with finding reference token sequences $r$ during each attribution map generation.

To demonstrate the effectiveness of the reference library, we conducted additional experiments comparing the attribution quality of Contrast-CAT utilizing the reference library (deterministic) versus Contrast-CAT using a constrained random sampling approach, where $30$ samples were randomly selected from the training set that satisfy the condition $f_c(r) < \gamma$ for each attribution map generation.
Notably, this constrained random sampling approach can be considered an on-the-fly version of Contrast-CAT.

Attribution quality was evaluated using faithfulness metrics (AOPC and LOdds) measured under the MoRF setting with the BERT$_{\text{base}}$ model on the SST2 dataset.
For the constrained random sampling approach, the experiment was repeated $30$ times, and the averaged results, along with their standard deviations, were reported. 
The results are presented in Table~\ref{tbl:perf_comparison_ref_lib}.

In addition to the attribution quality comparison, we also measured and compared the runtime of Contrast-CAT using the reference library (denoted as ``Reference Library" in Table~\ref{tbl:time_comparison_ref_lib}) with the on-the-fly version of Contrast-CAT (denoted as ``On-the-Fly" in Table~\ref{tbl:time_comparison_ref_lib}).
Table~\ref{tbl:time_comparison_ref_lib} summarizes the average runtime (in seconds) required to generate an attribution map for a single sample using each approach.

The results in Tables~\ref{tbl:perf_comparison_ref_lib} and~\ref{tbl:time_comparison_ref_lib} demonstrate the effectiveness of using the reference library in terms of both attribution quality and computational efficiency. 
Specifically, it achieves $\times 2.24$ faster runtime and slight improvements in AOPC and LOdds compared to the mean values of the on-the-fly version of Contrast-CAT.
It is also noteworthy that, while Contrast-CAT utilizing the reference library outperforms the on-the-fly version, the on-the-fly version still exhibits superior attribution quality and runtime compared to the second-best competing method, TIS.

\section{Impact of Deletion Threshold on Attribution Quality}\label{appnedix.rho}

In all experiments presented in this paper, we consistently used the mean plus one standard deviation (mean + std) as the deletion threshold $\rho$, as described in Section~\ref{sec:multi-references} (“Attribution with Multiple Contrast: Refinement via Deletion Test”).
To further examine how the choice of $\rho$ affects the performance of Contrast-CAT, we conducted an additional analysis comparing three settings: mean – std, mean, and mean + std.

Table~\ref{tbl:sensitivity_rho} reports AOPC and LOdds scores for Contrast-CAT under the MoRF setting (as in Table~\ref{tbl:all.measures}~(A)), evaluated on four datasets—Amazon, Yelp, SST2, and IMDB—using the pre-trained BERT$_\text{base}$ model. For comparison, we also include the performance of the TIS method, which ranked among the top two baselines in Table~\ref{tbl:all.measures}~(A) of the main paper.
\begin{table*}[tb]
\centering
\caption{AOPC and LOdds scores of Contrast-CAT under different deletion thresholds $\rho$ (mean – std, mean, mean + std), evaluated under the MoRF (Most Relevant First) setting using the pre-trained BERT$_{\text{base}}$ model.}\label{tbl:sensitivity_rho}
%\small
\newcolumntype{C}[1]{>{\centering\arraybackslash}m{#1}}
\begin{tabular}{c|c|c|c|c|c|c|c|c}\toprule
 Dataset& \multicolumn{2}{c|}{Amazon} & \multicolumn{2}{c|}{Yelp} & \multicolumn{2}{c|}{SST2}& \multicolumn{2}{c}{IMDB} \\\midrule%\cline{2-7}
 $\rho$ & AOPC$\uparrow$ & LOdds$\downarrow$ & AOPC$\uparrow$ & LOdds$\downarrow$ & AOPC$\uparrow$ & LOdds$\downarrow$ & AOPC$\uparrow$ & LOdds$\downarrow$ \\
\midrule
Mean - Std & 0.557 & 0.265 & 0.574 & 0.243 & 0.493 & 0.339 & 0.684 & 0.152 \\
Mean & 0.630 & 0.193 & 0.633 & 0.186 & 0.577 & 0.249 & 0.714 & 0.125 \\
Mean + Std & 0.703 & 0.117 & 0.687 & 0.131 & 0.654 & 0.157 & 0.738 & 0.101\\\hline
TIS (Baseline) & 0.560 & 0.241 & 0.494 & 0.349 & 0.463 & 0.367 & 0.618 & 0.277\\\bottomrule
\end{tabular}
\end{table*}

Overall, the results in Table~\ref{tbl:sensitivity_rho} suggest that using a conservative threshold (mean + std) yields more precise and faithful attribution maps.

\section{Examples of Reference Sentences}\label{appnedix.fig1_running_example}
Table~\ref{tbl:single_input_example} lists the five reference sentences used by Contrast-CAT to generate the attribution map in Figure~\ref{fig:motivation} for the input “It is very slow.”, which was predicted as negative. These reference sentences are drawn from the opposite sentiment class (positive) and yield low confidence scores for the target class (negative).
\begin{table}[!ht]
\centering
\caption{Reference sentences used by Contrast-CAT to interpret the input “it is very slow”, which is classified as negative by the pre-trained BERT$_{\text{base}}$ model. These references satisfy $f_c(r) < \gamma$ for the negative class $c$.}\label{tbl:single_input_example}
\setlength{\tabcolsep}{8pt}
\begin{tabular}{c}\toprule
Reference Sentence(s)  \\\bottomrule
It is a whole lot of fun \\
A delightful, witty, improbable romantic comedy with a zippy jazzy score. \\
Exhilarating, funny and fun. \\
This charming, thought-provoking new york fest of life and love. \\
Great fun, full of the kind of energy it's documenting. \\
\bottomrule
\end{tabular}
\end{table}

Table~\ref{tbl:ref_sets_by_class} lists the top-5 reference sentences $r$ for each target class $c$ in the SST2 dataset that satisfy $f_c(r) < \gamma$—i.e., sentences that yield low confidence scores for class $c$—as selected by Contrast-CAT. 

\begin{table}[!ht]
\centering
\caption{Top-5 reference sentences $r$ satisfying $f_c(r) < 0.001$ for each target class $c$ in the SST2 dataset. These references are used to construct contrastive attributions in Contrast-CAT.}\label{tbl:ref_sets_by_class}
\setlength{\tabcolsep}{8pt}
\begin{tabular}{c|c}\toprule
\multicolumn{2}{c}{Target Class $c$ : Positive}\\\bottomrule
$f_c(r)$ & Token sequence $r$  \\
\hline
\multirow{5}{*}{$< 0.001$} & Excruciatingly unfunny and pitifully unromantic \\
& Contains very few laughs and even less surprises \\ %0.00081353786
& It’s simply stupid, irrelevant and deeply flawed \\
& Scuttled by a plot that's just too boring and obvious \\
& Unlikable, uninteresting, unfunny, and completely, utterly inept\\
\midrule
\multicolumn{2}{c}{Target Class $c$ : Negative}\\\bottomrule
$f_c(r)$ & Token sequence $r$  \\
\hline
\multirow{5}{*}{$< 0.001$} & Very well written and very well acted \\
& A film of freshness, imagination and insight \\
& A wild, endearing, masterful documentary \\
& Delighted with the fast, funny, and even touching story\\
& A truly wonderful tale combined with stunning animation \\
\bottomrule
\end{tabular}
\end{table}

\section{Activation Visualization}\label{appnedix.act_visu}

\begin{figure}[!ht]
\centering
\includegraphics[width=0.95\textwidth]{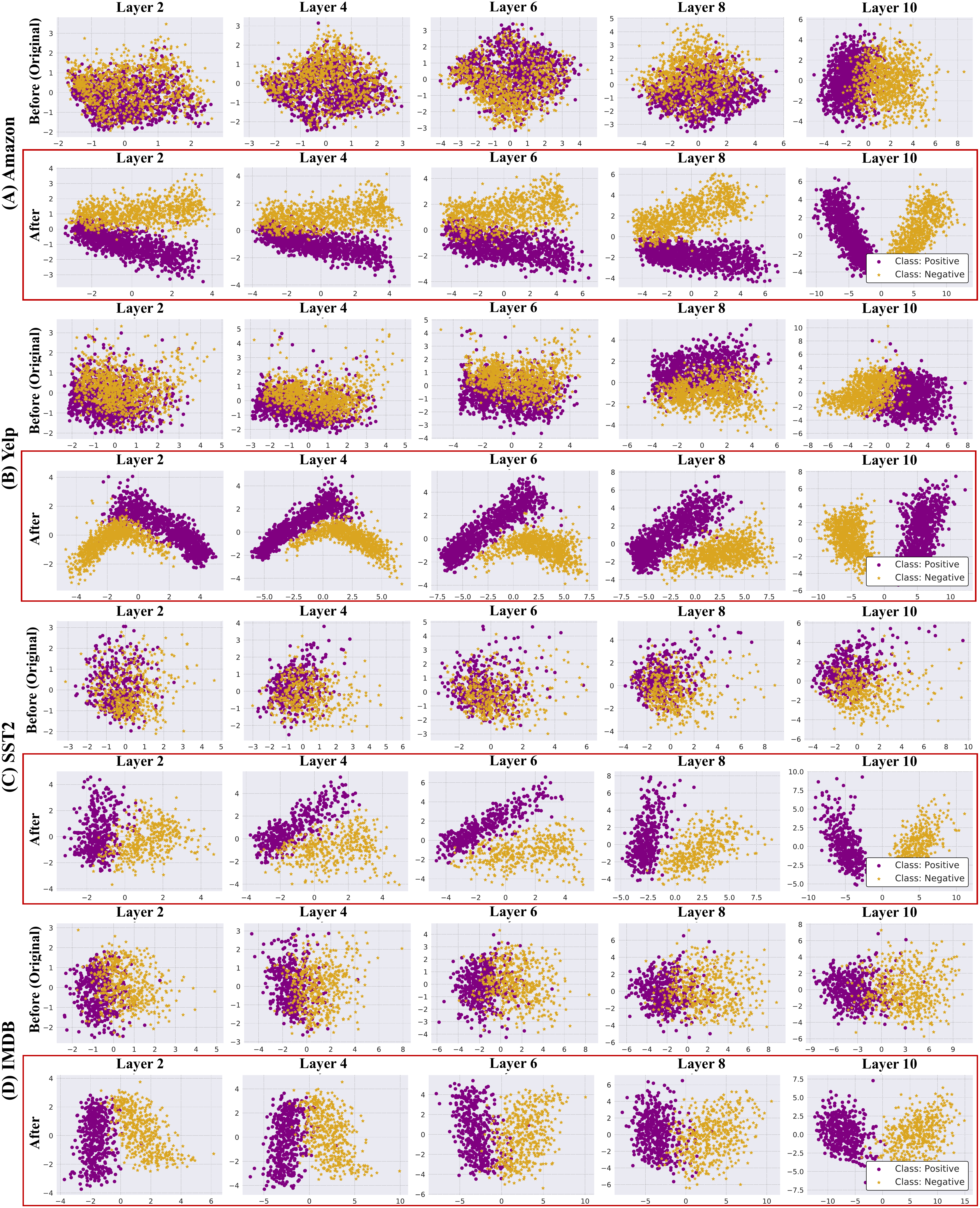}
\caption{Visual representation of activations across five different layers of the \textbf{BERT$_{\text{base}}$} model for four different datasets: (A) Amazon, (B) Yelp, (C) SST2, and (D) IMDB. Odd-numbered rows show activations before applying \mymethod's multiple contrasting, and even-numbered rows (highlighted in a red box) show activations after applying \mymethod's multiple contrasting. The colors represent classes: positive (\textcolor[rgb]{0.8627,0.6510,0.1137}{yellow}) and negative (\textcolor[rgb]{0.5059,0,0.5059}{purple}). Principal Component Analysis is used to reduce the dimensionality of activations to two dimensions for visualization.
The separation between positive (\textcolor[rgb]{0.8627,0.6510,0.1137}{yellow}) and negative (\textcolor[rgb]{0.5059,0,0.5059}{purple}) classes becomes more distinct after applying \mymethod's multiple contrasting.}
\label{fig:act_visu_bert}
\end{figure}

To demonstrate that \mymethod's multiple activation contrasting, detailed in Section~\ref{sec:multi-references}, effectively reduces non-target signals (class-irrelevant features) within target activation, we visualized activations from different layers of the BERT$_{\text{base}}$ models, as shown in Figure~\ref{fig:act_visu_bert}. The 1st, 3rd, 5th, and 7th rows (odd-numbered rows) represent the original activations, while the 2nd, 4th, 6th, and 8th rows (even-numbered rows) show the activations after applying \mymethod's multiple activation contrasting.
Each point represents the averaged activation across tokens, extracted from the corresponding layers.
For visualization, the dimensionality of these averaged activations was reduced to two using Principal Component Analysis~\citep{pca}.

As illustrated in Figure~\ref{fig:act_visu_bert}, the original activations (odd-numbered rows in the figure) show poor separation between positive and negative classes.
In contrast, after applying \mymethod's multiple activation contrasting (even-numbered rows in the figure), the activations exhibit much clearer class separation across all layers.
This enhanced separation highlights the effectiveness of \mymethod in reducing class-irrelevant features within activations, thereby improving attribution quality by focusing on class-relevant features.

\section{Extending Contrast-CAT Beyond Text Classification}\label{appnedix.future_direction}
While our work focuses on interpreting transformer-based models for text classification, Contrast-CAT has the potential to generalize beyond this setting. 

Because Contrast-CAT operates on token-level activations and attention patterns, it is not inherently constrained to text classification. Many other transformer-based tasks—such as question answering and natural language inference—also rely on token-level representations and could similarly benefit from contrastive attribution. Moreover, in domains beyond NLP, such as computer vision and multimodal learning, transformer architectures like Vision Transformers (ViTs) and text-image models also process token-like representations in a tokenized form. These architectures present promising opportunities to extend Contrast-CAT for interpreting model decisions in non-textual settings. We also foresee its applicability in areas such as transformer-based source code vulnerability detection~\citep{sim_vul_sm}, where explainability is critical for enhancing model reliability and trust.

However, applying Contrast-CAT to new tasks or modalities will likely require task-specific adaptations. In particular, identifying suitable reference activations is non-trivial, as it depends on the semantic and structural characteristics of each modality. Future work could explore novel strategies for reference selection and contrast construction tailored to diverse application contexts.

We view this study as a foundational step toward broader, cross-domain interpretability of transformer-based models, paving the way for safer and more transparent AI systems across a wide range of applications.

\end{document}